\pdfoutput=1

\documentclass[11pt]{article}

\usepackage[]{emnlp2022}

\usepackage{times}
\usepackage{latexsym}

\usepackage[T1]{fontenc}

\usepackage[utf8]{inputenc}

\usepackage{microtype}

\usepackage{amssymb}
\usepackage{algorithm}
\usepackage{algpseudocode}
\usepackage{booktabs}
\usepackage{graphicx}
\usepackage{subfig}
\usepackage{caption}
\usepackage[normalem]{ulem}

\usepackage{amsmath}
\usepackage{tablefootnote}
\usepackage{mathtools}
\usepackage{enumitem}

\newcommand\eg{\textit{e.g.}}
\newcommand\ie{\textit{i.e.}}

\title{Discovering Language-neutral Sub-networks \\ in Multilingual Language Models}

\author{Negar Foroutan    \qquad Mohammadreza Banaei  \qquad  R\'emi Lebret \\ {\bf Antoine Bosselut}  \qquad {\bf Karl Aberer} \\
    \texttt{\normalsize \{firstname.lastname\}@epfl.ch}
      \\ EPFL}

\definecolor{tropicalrainforest}{rgb}{0.0, 0.46, 0.37}

\begin{document}
\maketitle
\begin{abstract}

Multilingual pre-trained language models transfer remarkably well on cross-lingual downstream tasks.
However, the extent to which they learn language-neutral representations (\ie, shared representations that encode similar phenomena across languages), and the effect of such representations on cross-lingual transfer performance, remain open questions.

In this work, we conceptualize language neutrality of multilingual models as a function of the overlap between language-encoding sub-networks of these models.
We employ the lottery ticket hypothesis to discover sub-networks that are individually optimized for various languages and tasks.
Our evaluation across three distinct tasks and eleven typologically-diverse languages demonstrates that sub-networks for different languages are topologically similar (\ie, language-neutral), making them effective initializations for cross-lingual transfer with limited performance degradation.\footnote{Our code is available at \href{https://github.com/negar-foroutan/multiLMs-lang-neutral-subnets}{https://github.com/negar-foroutan/multiLMs-lang-neutral-subnets}.}

\end{abstract}

\section{Introduction}
\label{sec:introduction}


Multilingual language models~(MultiLMs) such as mBERT~\cite{devlin2018bert}, XLM~\cite{conneau2019cross}, and XLM-R~\cite{conneau2019unsupervised} 
are pre-trained jointly on raw data from multiple languages.
\begin{figure}[!ht]
\centering
\subfloat[Step 1: Discover sub-networks in multilingual language models that encode particular languages]{
        \label{fig:methodology_step1}
        {\includegraphics[width=0.79\columnwidth]{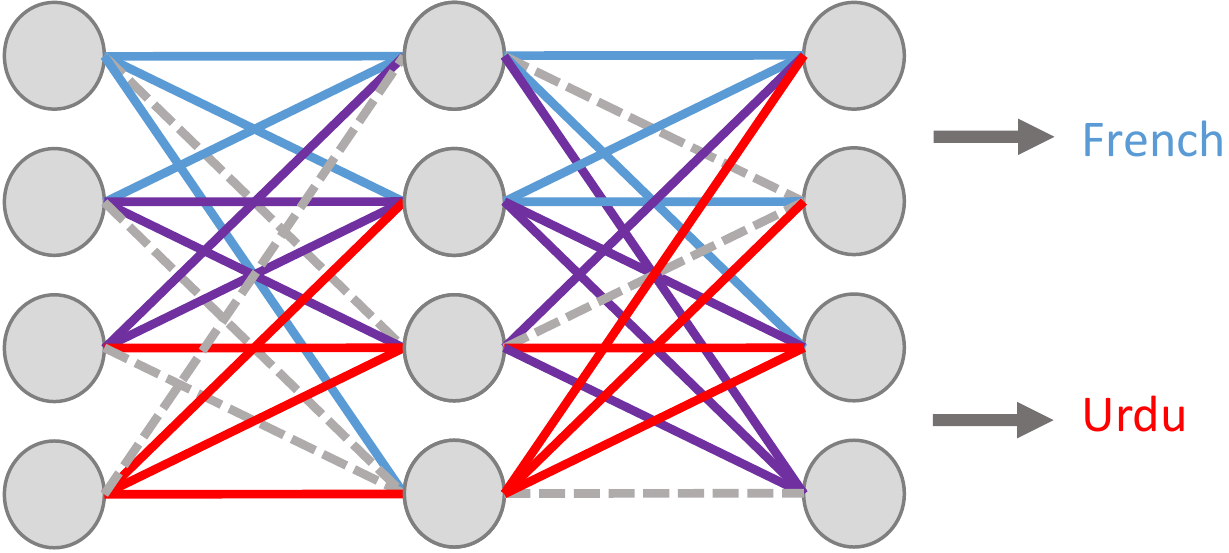}}
    }

\vspace{-2mm}

\subfloat[Step 2: Transfer sub-networks to new languages to evaluate their language neutrality degree]{
        \label{fig:methodology_step2}
        {\includegraphics[width=0.79\columnwidth]{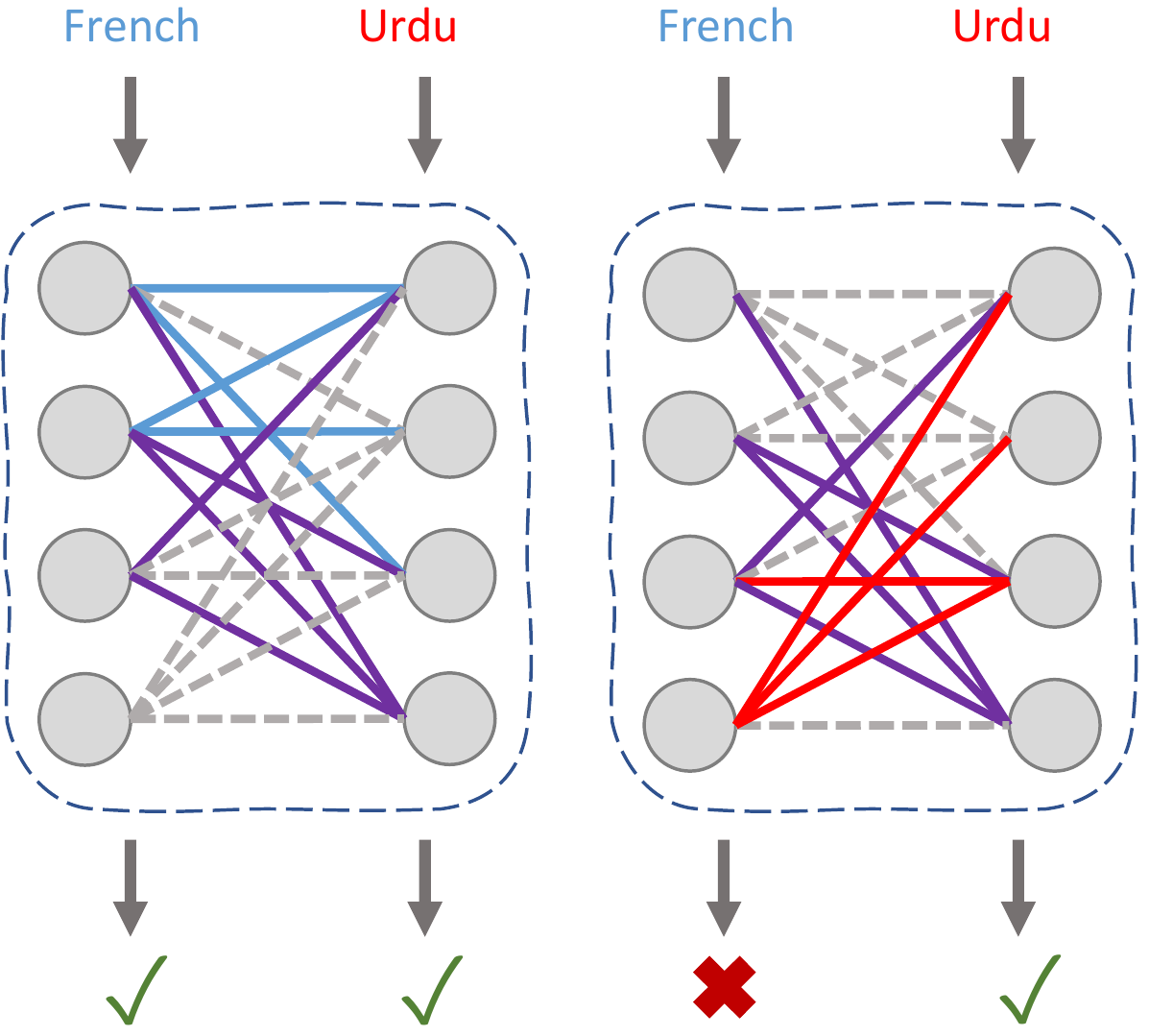}}
    }
\caption{Overview of our approach.
We discover sub-networks in the original multilingual language model that are good foundations for learning various tasks and languages (a). Then, we investigate to what extent these sub-networks are similar by transferring them across other task-language pairs (b).
In this example, the \textcolor{blue}{blue} and \textcolor{red}{red} lines show sub-networks found for French and Urdu, respectively, and \textcolor{violet}{purple} connections are shared in both sub-networks. Dashed lines show the weights that are removed in the pruning phase. }
\label{fig:method_visualization}
\vspace{-6mm}
\end{figure}
Later, when they are fine-tuned for a task using perhaps a single high-resource language dataset, they demonstrate promising zero-shot cross-lingual performance, generalizing to the same task in different languages despite not having been fine-tuned for those languages.

The facilitator of this cross-lingual transfer ability is often hypothesized to be a learned language neutrality in these models (\ie, similar linguistic phenomena across languages are represented similarly by the model). However, the source of this ability remains an open question. Certain studies have investigated mBERT~(a prominent MultiLM) 
for universal language-neutral components that would facilitate cross-lingual transfer~\cite{libovic2019how, pires2019multilingual, libovicky2020language}. Meanwhile, other studies claim mBERT is not language-neutral as it partitions its multilingual semantic space into separate language-specific subspaces~\cite{singh2019bert, choenni2020does, wu2019beto}.
However, most of these prior works analyze the language neutrality of MultiLMs by probing their output contextual representations.

In this paper, language-neutrality is instead conceptualized in terms of the parameters of MultiLMs. We hypothesize that a multilingual model with language-neutral representations would have learned different languages using the same subset of parameters in its network structure. 
Sub-networks in MultiLMs that overlap across languages, and transfer well when re-trained on other languages from the pre-training corpus, would indicate that the model is comprised of language-neutral representations that jointly encode multiple languages.
%
%
To demonstrate this effect, we extract sub-networks from MultiLMs by pruning them for individual languages and task pairs using iterative magnitude pruning~\citep{frankle2018lottery}~(Figure~\ref{fig:methodology_step1}). 
Then, 
we evaluate the cross-lingual transfer of sub-networks by re-initializing to the unpruned parameters of the original MultiLM, and re-training on task data for different languages~(Figure~\ref{fig:methodology_step2}). 

Our results on three tasks, namely masked language modeling, named entity recognition~\citep{pan2017cross}, and natural language inference~\citep{conneau2018xnli}, show high absolute parameter overlap among sub-networks discovered for different languages and effective cross-lingual transfer between languages on the same task (even outperforming the original multilingual model for certain low-resource languages). 
However, cross-lingual sub-network transfer deteriorates as we increase the sparsity level of the sub-networks, suggesting that language-neutral components of the MultiLMs are pruned in favor of retaining necessary language- and task-specific components for the language-task pair used to discover the sub-network.

\section{Background \& Motivation}
\label{sec:related_work}

\paragraph{Multilingual Language Models.}

Multilingual language models~(MultiLMs) such as mBERT~\cite{devlin2018bert}, XLM~\cite{conneau2019cross}, and XLM-R~\cite{conneau2019unsupervised} have achieved state of the art results in cross-lingual understanding tasks by jointly pre-training Transformer models~\cite{vaswani2017attention} on many languages.
Specifically, mBERT has shown effective cross-lingual transfer for many tasks, including named entity recognition~\cite{pires2019multilingual, wu2019beto}, cross-lingual natural language inference~\cite{conneau2018xnli, junjie2021xtreme}, and question answering~\cite{patrick20QA}.

Due to these impressive cross-lingual transfer results, many recent works investigate the source of this capacity.
One line of study investigates the effect of different pre-training settings (\eg, shared vocabulary, shared parameters, joint multilingual pre-training, etc.) on cross-lingual transfer \citep{wang2019cross, artetxe2020cross, conneau2020emerging}.
%
Other works explore how learned multilingual representations are partitioned into language-specific subspaces \citep{singh2019bert, wu2019beto, choenni2020does}. 
%
In contrast to these works, our study explores the existence of language-neutral \textit{parameters} in MultiLMs. 
Using iterative magnitude pruning, we extract pruned sub-networks from multilingual LMs for various tasks and languages, and investigate to what extent these sub-networks are similar by transferring them across languages.

\paragraph{Analysis of Multilingual Representations.}

Prior work has demonstrated the lack of language neutrality in mBERT by comparing the similarity of mBERT's encodings for semantically-similar sentences across multiple languages~\cite{singh2019bert}, concluding that mBERT partitions the representation space among languages rather than using a shared, interlingual space.\footnote{In Appendix~\ref{sec:cca_representation}, we discuss how the choice of the similarity metric can affect the results of such analysis.} 
In opposition, other efforts show that mBERT learns language-neutral representation space that facilitates cross-lingual transfer~\cite{pires2019multilingual, libovic2019how, libovicky2020language}.
These findings have led to attempts to disentangle the language-specific and language-neutral components of mBERT to improve its performance~\cite{libovic2019how, libovicky2020language, Zirui2020negativeinterference, gonen2021greek, zhao2021inducing, zehui2021subnetMMT}.
The language-neutral component itself can be viewed as the stacking of two sub-networks: a multilingual encoder followed by a task-specific language-agnostic predictor~\cite{muller2021firstalign}.

In this work, we discover language-neutral components in the parameters of MultiLMs by pruning them for different language-task pairs.
By transferring sub-networks across languages, we investigate to what extent these sub-networks are language-neutral.

\paragraph{Lottery Ticket Hypothesis.}

The lottery ticket hypothesis~(LTH) shows that dense, randomly-initialized neural networks contain small, sparse sub-networks~(\ie, winning tickets) capable of being trained in isolation to reach the accuracy of the original network~\cite{frankle2018lottery}.
This phenomenon has been observed in multiple applications, including computer vision \citep{morcos2019oneticket, frankle2020linear} and natural language processing \cite{gale2019state, yu2019playing}.

In particular, the lottery ticket hypothesis has previously been applied to the BERT model \citep{devlin2018bert} using the GLUE benchmark~\cite{wang2019glue, prasanna2020allwinning, chen2020lottery}.
Importantly, \citet{chen2020lottery} show that sub-networks found using the masked language modeling task can transfer to other tasks. 
In the context of multilingual language models, \citep{ansell2022composable} propose a sparse LTH-based fine-tuning method to benefit from both modular and expressive fine-tuning approaches. 
In this work, we specifically use the lottery ticket hypothesis to examine the possibility of transferring winning tickets across languages, thereby disentangling whether language-neutral and language-specific components of mBERT emerge in winning tickets of different languages.
\section{Methodology}
\label{sec:methodology}
In this section, we define sub-networks (\ie, winning tickets in the LTH) and identify our approach to discovering them through a pruning algorithm derived from the lottery ticket hypothesis.

\paragraph{Sub-networks.}
For any network $f(x;\theta)$ with initial parameters $\theta$, we define a sub-network for $f$ as $f(x; m \odot \theta)$ where $m \in \{0, 1\}^{|\theta|}$ is a binary mask on its parameters and $\odot$ is the element-wise product. 
The mask $m$ sets many of the parameters of the original network to zero, which prunes edges in the computation graph and yields a sub-network of the original network.

\paragraph{Winning ticket.}
When training network $f(x;\theta)$ on a task, $f$ reaches maximum evaluation performance $a$ at iteration $i$.
Similarly, when training sub-network $f(x; m \odot \theta)$ on the same task, $f$ reaches maximum evaluation performance $a^\prime$ at iteration $i^\prime$. A sub-network $f(x; m \odot \theta)$ is a winning ticket $f^*(x; m \odot \theta)$ if it achieves similar or better performance than the original network: $a - a' \leq \epsilon$ when $i^\prime \leq i$ (we set $\epsilon$ to be one standard deviation of  performance of the original mBERT model). 

\paragraph{Identifying winning tickets.}
As shown in Algorithm~\ref{alg:mip}, we use unstructured magnitude iterative pruning~\cite{han2015learning} to discover winning tickets.
In this approach, an original model is repeatedly trained on a task over multiple rounds $r$, and a subset of its parameters is pruned in each round until a desired sparsity level $s$ is reached~\cite{frankle2018lottery, chen2020lottery}.
More specifically, in each round $r$ of training, the model is trained to its peak performance on the task.
Then we prune $p\%$ of the original parameters with the lowest magnitudes.
After pruning the subset of the weights in a round $r$, the remaining weights are reset to their original pre-trained initialization, and the model is re-trained on the same dataset again. We set the iterative pruning rate $p=10$ in all experiments (e.g., five rounds to reach $s=50\%$ sparsity). To evaluate whether a sub-network is a winning ticket for a task, we check the network's task-specific performance once it has reached the desired sparsity level $s$.
Figure~\ref{fig:method_visualization} shows an overview of our approach.

\begin{algorithm}[t]
\footnotesize
\caption{Iterative magnitude pruning to reach sparsity level $s$.}
\label{alg:mip}

\begin{algorithmic}
\State Start with mBERT pre-trained weights $\theta_0$
\State Set $p \gets 10$ and $n \gets \frac{s}{p}$
\For{$r \gets 0$ to $n-1$}                 
    \State Train network for $i$ iterations, arriving at parameters $\theta_i$
    \State Prune $p\%$ of the parameters in $\theta_0$
    \State Reset the remaining parameters to their values in $\theta_0$
\EndFor
\State Return the resulted pruned network
\end{algorithmic}
\end{algorithm}

\begin{figure*}
    \centering
    \subfloat{\includegraphics[width=0.55\textwidth]{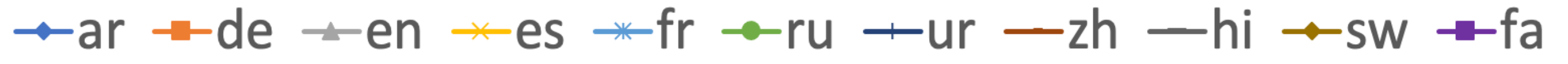}}
    
    \addtocounter{subfigure}{-1}
    
    \centering
    \subfloat[MLM]{
        \label{fig:mlm_sparsity}
        {\includegraphics[width=0.32\textwidth]{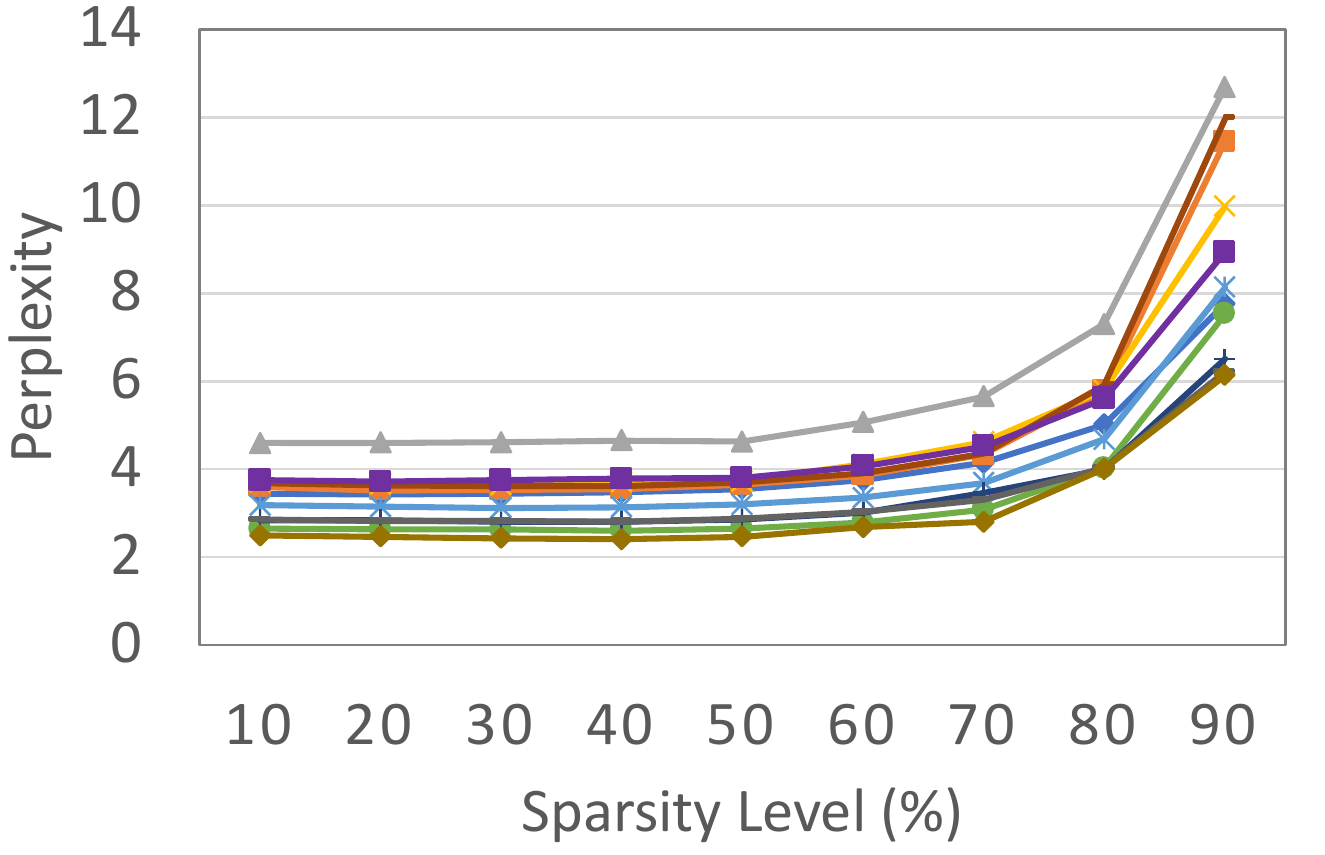}}
    }
    \centering
    \subfloat[NER]{
        \label{fig:ner_sparsity}
        {\includegraphics[width=0.32\textwidth]{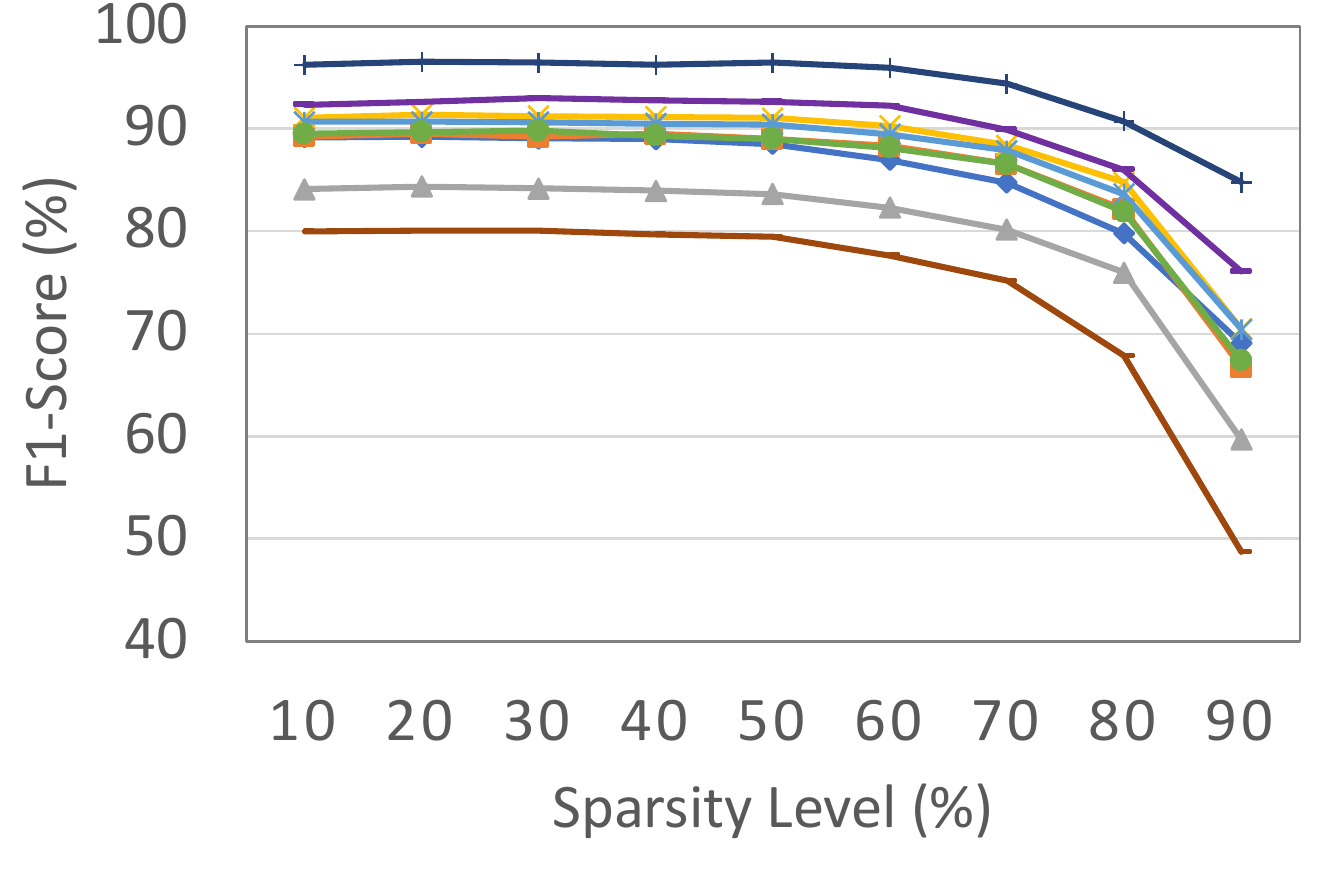}}
    }
    \centering
    \subfloat[XNLI]{
        \label{fig:xnli_sparsity}
        {\includegraphics[width=0.32\textwidth]{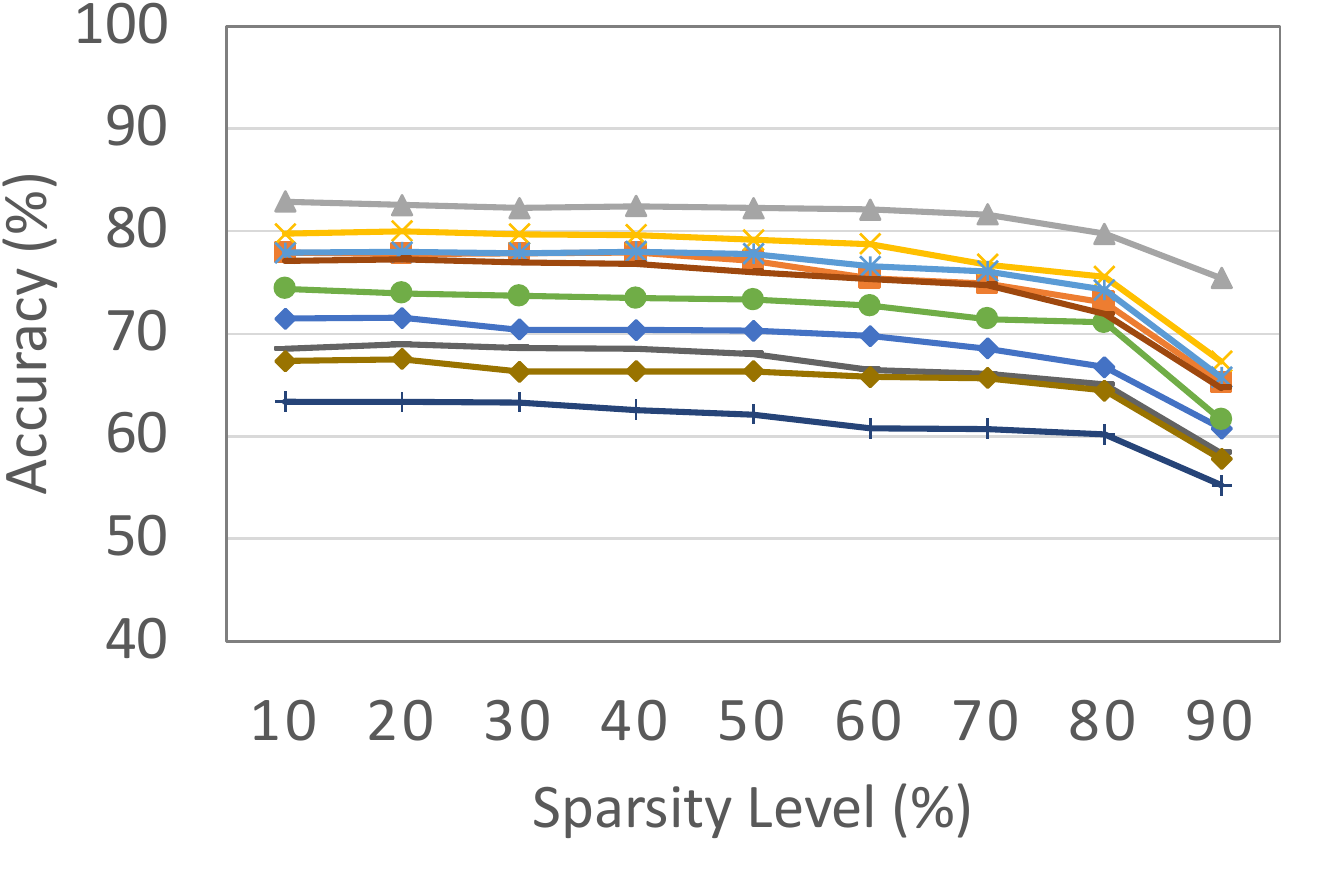}}
    }
    \caption{Performance of pruned sub-networks for each task-language pair at different sparsity levels}
    \label{fig:sparsity_performance}
    \vspace{-4mm}
\end{figure*}

\section{Experimental Setup}
\label{sec:experimental_setup}
\subsection{Model}

In our experiments, we use the cased version of multilingual BERT (mBERT; \citealp{devlin2018bert}).\footnote{\href{https://huggingface.co/bert-base-multilingual-cased}{https://huggingface.co/bert-base-multilingual-cased}}
This model is a 12-layer transformer with around 110M parameters. It is pre-trained using the masked language modeling and next sentence prediction training objectives on the Wikipedia dumps for 104 languages. A shared Wordpiece \citep{wu2016wordpiece} vocabulary of size 110k is initialized for these 104 languages.

\subsection{Tasks \& Datasets}

We perform our experiments on three different NLP tasks.
We have chosen typologically diverse languages covering different language families: Germanic, Romance, Indo-Aryan, and Semitic, and including both high- and low-resource languages from the NLP perspective. 
These languages are including English~(en), French~(fr), German~(de), Chinese~(zh), Russian~(ru), Spanish~(es), Farsi~(fa), Urdu~(ur), Arabic~(ar), Hindi~(hi), and Swahili~(sw).

\paragraph{Masked Language Modeling (MLM).} For this task, we use 512-token sequences from Wikipedia in each language as the training data.
    
\paragraph{Named Entity Recognition (NER).} We use WikiAnn~\cite{pan2017cross}, a multilingual named entity recognition and linking dataset built on Wikipedia articles for 282 languages\footnote{\href{https://huggingface.co/datasets/wikiann}{https://huggingface.co/datasets/wikiann}}.
Swahili~(sw) is not included in this dataset, and Hindi~(hi) has a small data size, so these two languages are excluded from the NER experiments.
    
\paragraph{Natural Language Inference~(NLI).} We use the cross-lingual natural language inference (XNLI) dataset~\cite{conneau2018xnli}. This dataset includes a subset of examples from MNLI~\cite{N18-1101} translated into 14 languages\footnote{\href{https://huggingface.co/datasets/xnli}{https://huggingface.co/datasets/xnli}}.
Farsi is not included in this dataset.

\subsection{Training Details}

We tune hyperparameters and select evaluation metrics for each task based on prior work~\cite{chen2020lottery}\footnote{\href{https://github.com/VITA-Group/BERT-Tickets}{https://github.com/VITA-Group/BERT-Tickets}}.
All results are reported on the development sets of these datasets and are the average of three runs using different random seeds.
We use the same number of training and evaluation examples for all languages of a task.
Table~\ref{table:train_details} in the appendix reports pre-training and fine-tuning details. 
Computational details are reported in Appendix~\ref{sec:computation}.



\begin{figure*}

    \centering
    \subfloat{
        {\includegraphics[width=1\textwidth]{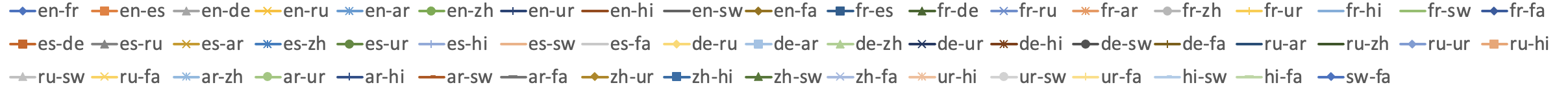}}
    }
    
    \addtocounter{subfigure}{-1}
    
    \centering
    \subfloat[MLM]{
        \label{fig:mlm_sparsity_overlap}
        {\includegraphics[width=0.32\textwidth]{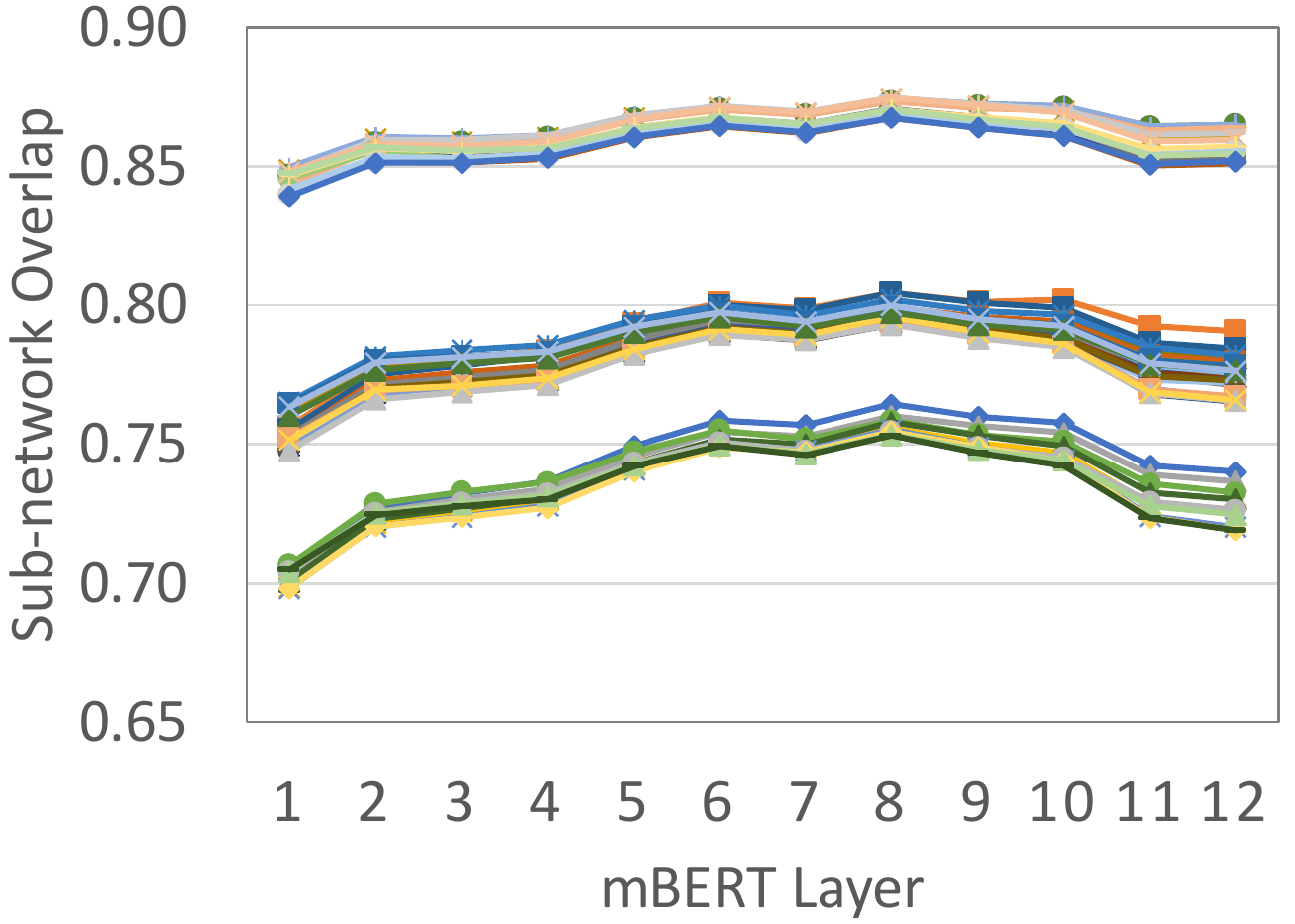}}
    }
    \centering
    \subfloat[NER]{
        \label{fig:ner_sparsity_overlap}
        {\includegraphics[width=0.32\textwidth]{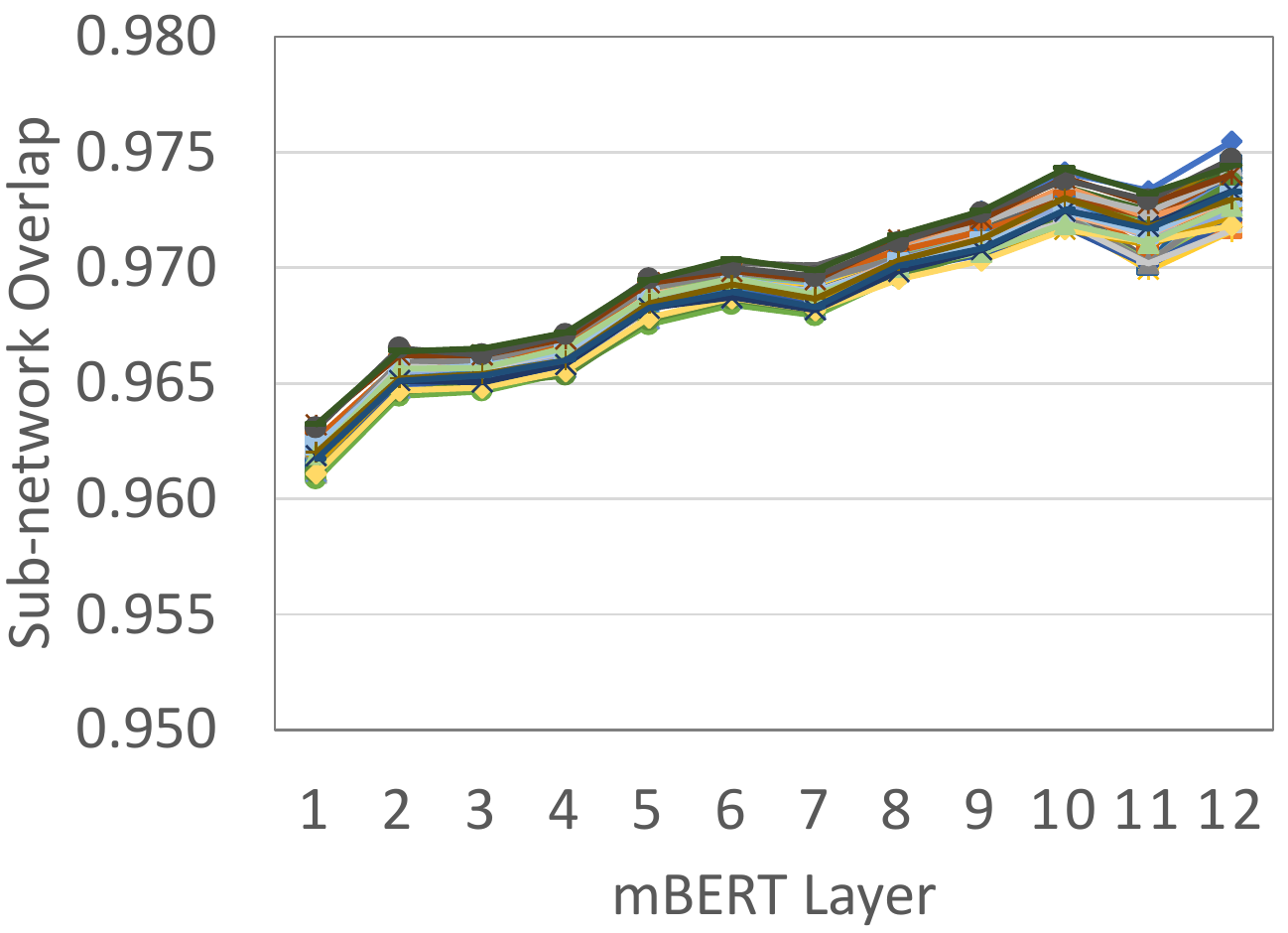}}
    }
    \centering
    \subfloat[XNLI]{
        \label{fig:xnli_sparsity_overlap}
        {\includegraphics[width=0.32\textwidth]{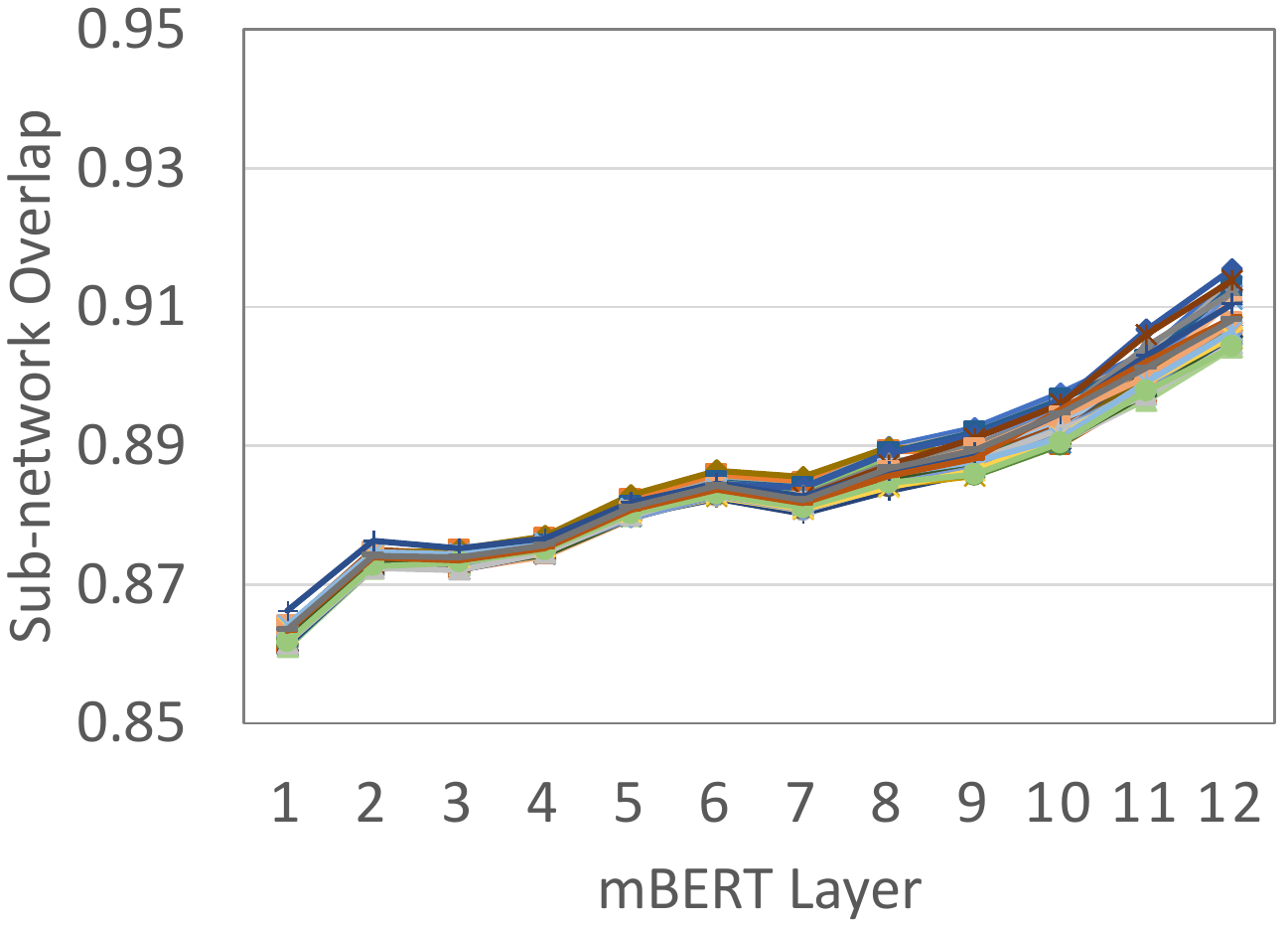}}
    }
    \caption{Sparsity pattern overlap between pruned sub-networks across different layers at 50\% sparsity level.}
    \label{fig:sparsity_overlap}
    \vspace{-4mm}
\end{figure*}

\section{Experimental Results}
\label{sec:results}

In this section, we report our evaluations to study the presence of language-neutral sub-networks in multilingual language models.
First, we validate the lottery ticket hypothesis for the mBERT model for various languages and tasks.
Then, we compare the winning tickets for different languages by analyzing their parameter overlap and cross-lingual performance.
Finally, to further assess the similarity of the winning tickets in a zero-shot setting, we evaluate their performance in a sentence retrieval task. 

\paragraph{Existence of Winning Tickets.}

To discover winning tickets for each task-language pair, we run Algorithm~\ref{alg:mip} to identify sub-networks at various sparsity levels. Then, we train the pruned sub-networks on the same task and language to identify whether it qualifies as a winning ticket, $f^*(x, m \odot \theta)$.

%
%
In Figure~\ref{fig:sparsity_performance}, we report the performance of pruned sub-networks for each task-language pair at different sparsity levels between 0 and 90\%. Winning tickets are found for each language and task at multiple sparsity levels, though as these sub-networks get sparser, they no longer satisfy the winning ticket criterion (performance must be within 1 standard deviation of the full mBERT performance). However, they still often perform within 10\% of their winning ticket performance. Interestingly, we find that the performance drop at higher sparsity levels is greater for the MLM and NER tasks than for XNLI, highlighting the importance of evaluating on tasks that induce diverse behavior in sparse models. 


\paragraph{Absolute Sub-network Overlap.}

Given that we can discover suitable winning tickets for every language and task, we now assess the similarity of the discovered sub-networks by comparing their parameter overlap.
Sub-network pairs (or larger groups) with higher overlap will be more likely to correspond to language-neutral representations in the original mBERT model. 
To measure parameter overlap between discovered sub-networks for different task-language pairs, we compute the Jaccard similarity between masks~($m_i$, $m_j$) from two sub-networks~$\big(\frac{m_i \cap m_j}{m_i \cup m_j}\big)$.
In Figure~\ref{fig:sparsity_overlap}, we report the parameter overlap across mBERT's layers for different language pairs across each of the tasks. These results are for sub-networks at $50\%$ sparsity, so we note that the expected Jaccard similarity of two randomly sampled $s$=$50\%$ sparse sub-networks would be $33\%$.

We find that the NER task generally discovers sub-networks with much more overlap across languages, followed by XNLI and MLM. Both NER and XNLI exhibit increasingly overlapping sub-networks at higher layers, while sub-networks discovered for the MLM task increase in overlap up until the middle layers and then drop again. This phenomenon is perhaps partly explained by the upper layers being more task-focused \citep{merchant2020whathappens}. For the MLM task, the model must predict language-specific tokens as the task. While we do not prune the task heads as part of our algorithm, the upper layers of mBERT may still need to be specialized to particular languages to predict their unique vocabularies\footnote{Further analysis of these MLM results is in Appendix~\ref{sec:subnet_overlap}.}.

As a comparison, we establish an upper bound on the expected overlap between languages, by computing the overlap of sub-networks for the same language~(pruned using different random seeds).
The average overlap of sub-networks across multiple runs is 98.66, 92.23, and 92.23 for the NER, XNLI, and MLM tasks, respectively.
In all three tasks, the overlaps across runs are much higher than the mask overlap across languages.

\paragraph{Cross-lingual Transfer of Winning Tickets.}
We now evaluate the transfer of winning tickets across languages for a given task. 
Using the discovered sub-networks for each task-language pair (set at a sparsity level of $50\%$ to maintain consistency across languages), we train these sub-networks on the other task-language pairs for the same task and compare their performance against the original sub-network trained on identical data (\ie, $a(s, t)$ vs $a(t, t)$\footnote{$a$ stands for the performance metric of the model as mentioned in Section~\ref{sec:methodology}.} where $s$ stands for the \textit{source} language and $t$ stands for the \textit{target} language).
%
If a transferred sub-network's performance on the \textit{target} language is within one standard deviation of the \textit{target} language sub-network's performance, we identify it as a successful transfer, indicating that this winning ticket is more language-neutral than language-specific, as its original parameters are adaptable for different languages. 

\begin{figure}
    \centering
    \subfloat[MLM]{
        \label{fig:MLM_transfer}
        {\includegraphics[width=0.86\columnwidth]{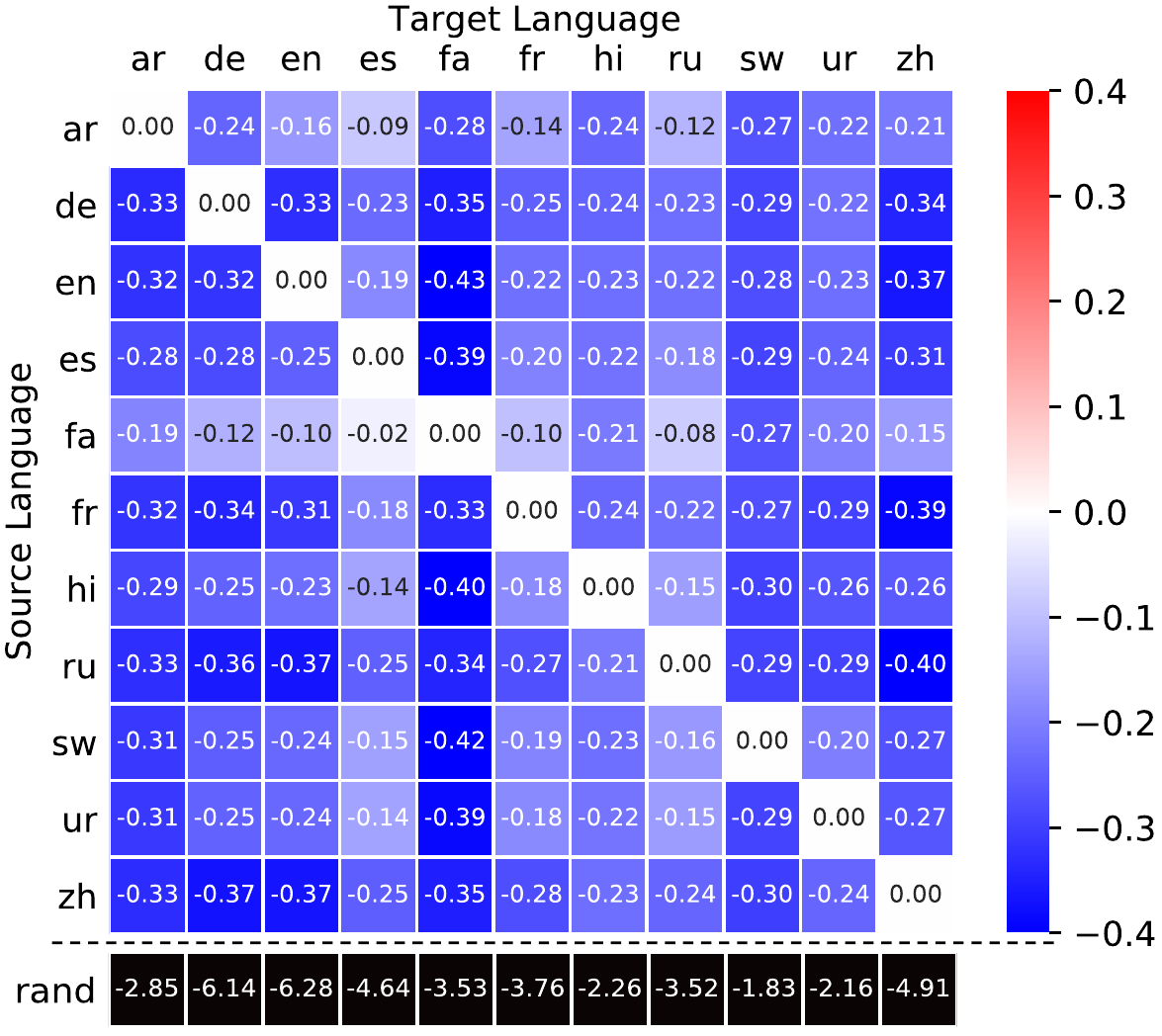}}
    }
    
    \subfloat[NER]{
        \label{fig:NER_transfer}
        {\includegraphics[width=0.86\columnwidth]{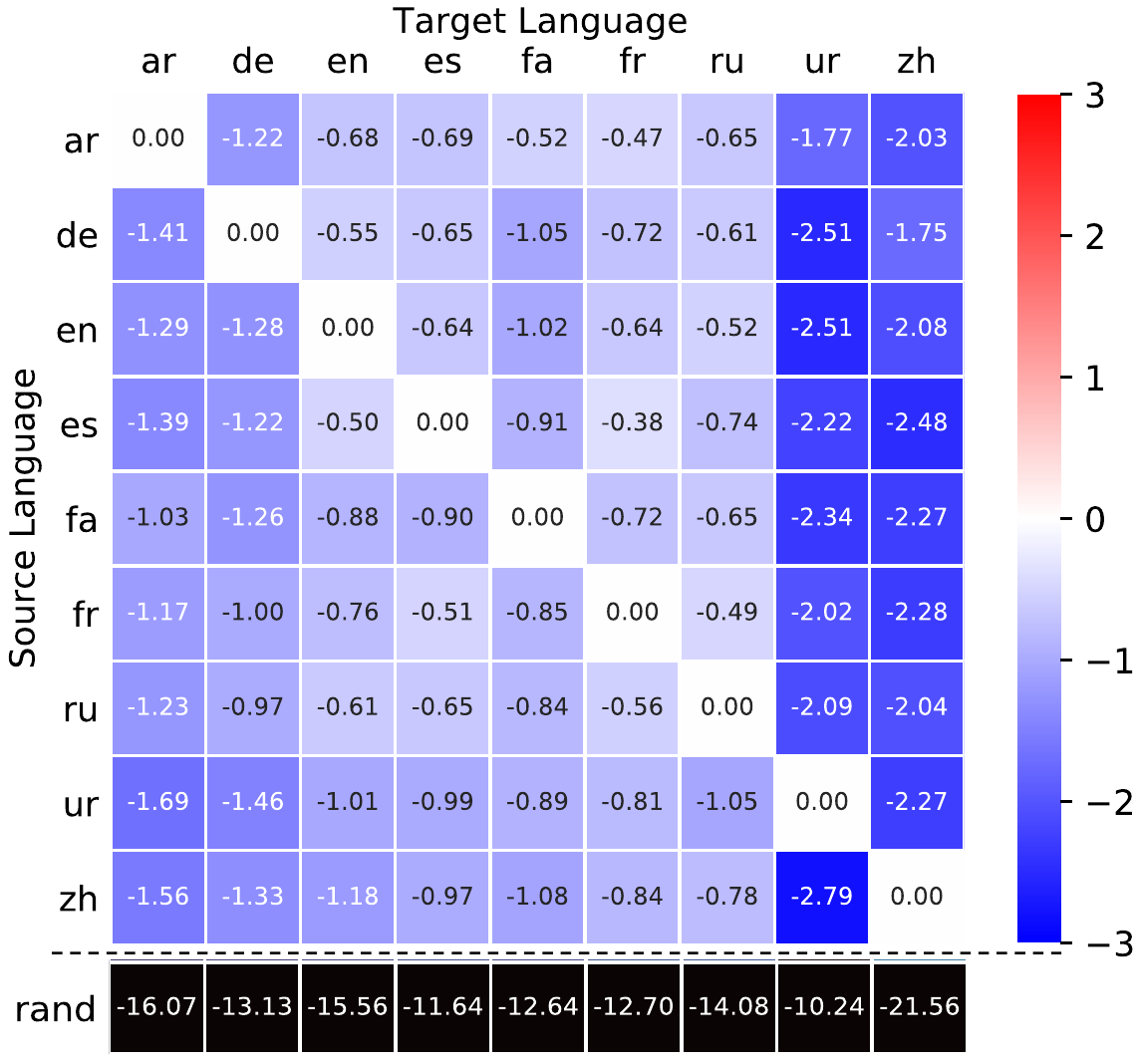}}
    }
    
    \subfloat[XNLI]{
        \label{fig:XNLI_transfer}
        {\includegraphics[width=0.86\columnwidth]{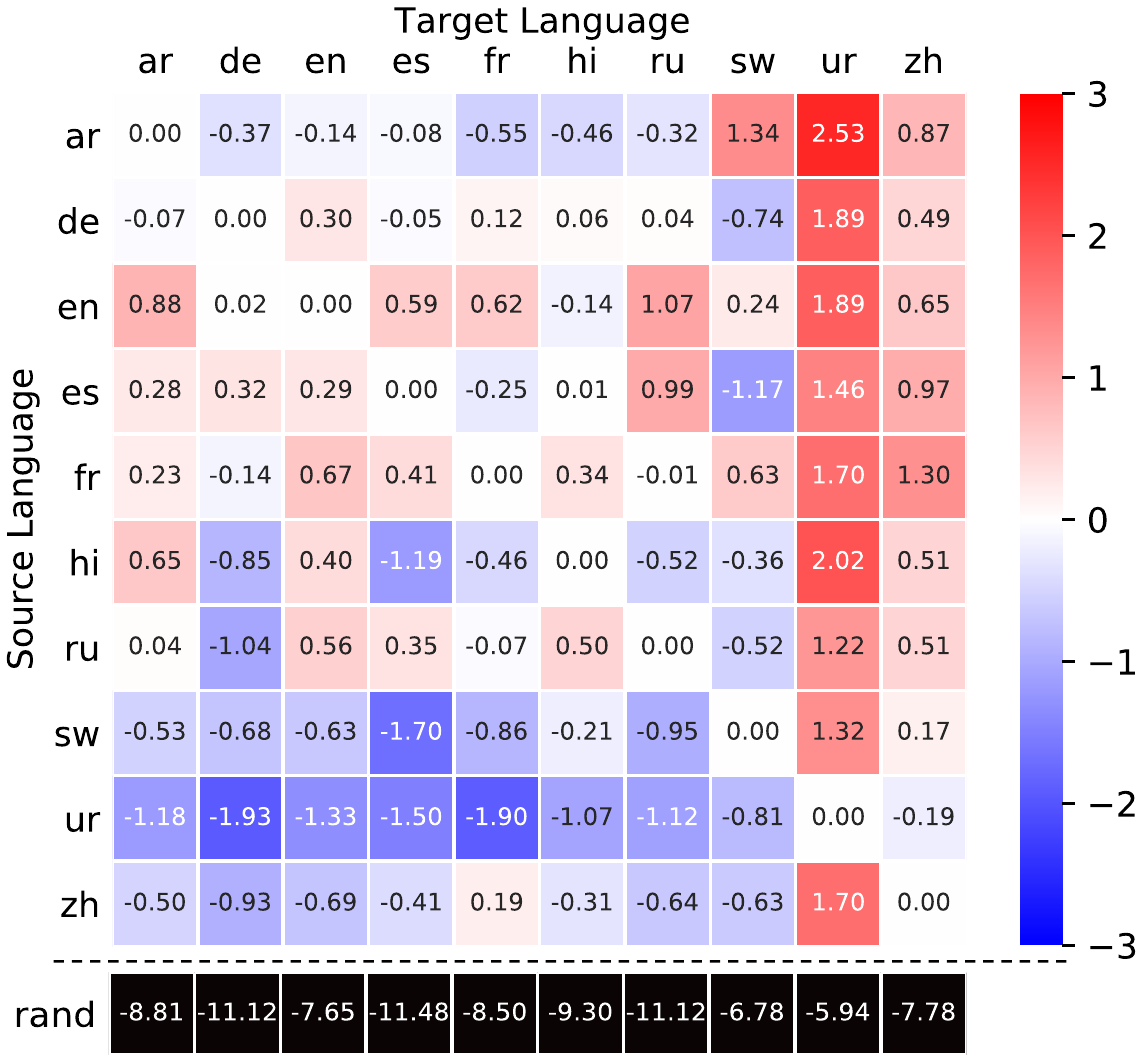}}
    }

    \caption{The performance difference of transferring winning tickets across languages~(50\% sparsity).
    Each row indicates the source language and each column indicates a target language.
    Each cell shows the difference in performance between the transferred sub-network and the sub-network discovered from training directly on the target language.
    We also report the performance drop of a random sub-network (\textbf{rand}) compared to the language sub-network.
    Blue and red colors indicate performance losses and gains respectively.
    }
    \label{fig:all_transfer}
    \vspace{-4mm}
\end{figure}

Figure~\ref{fig:MLM_transfer} shows the transfer performance (\ie, the performance drop or gain of the sub-network when trained on the \textit{target} language compared to when trained on the \textit{source} language) of MLM winning tickets. 
For this task, none of the winning tickets for source languages are winning tickets for other languages, meaning that the sub-networks do not achieve similar perplexity as to the \textit{target} sub-networks when transferred to the target tasks. 
However, the perplexity increase is relatively limited~(within $0.2$-$0.5$ points) compared to a $50\%$ sub-network pruned randomly (rand), suggesting that these language-specific sub-networks do contain shared multi-lingual components.

A similar transfer pattern emerges for the NER task in Figure~\ref{fig:NER_transfer}.
None of the transferred sub-networks match the performance of the source language sub-network. However, the performance remains high, with most target languages maintaining 98\% of the target sub-network performance when trained on the source sub-networks (compared to when trained on a random sub-network)\footnote{We also developed additional random baselines where LTH-discovered sub-networks at 10, 20, 30, and 40\% sparsity are randomly pruned to reach a 50\% sparse sub-network. Although the performance of these baselines is higher, transferring sub-networks across languages still outperforms them.}.
%
Chinese~(zh) and Urdu~(ur) experience the worst performance drop when we use sub-networks trained for other languages to transfer for these two languages.
One explanation could be that Chinese NER is more challenging than other languages due to the lack of capitalization information and the challenge of word segmentation in Chinese. Consequently, Chinese may require more language-specific information that may be pruned from the sub-networks of other languages.
The relatively low transfer performance for Urdu may be more empirical. Urdu's sub-network at 50\% sparsity has a high performance, outperforming mBERT by 1.2 points, making it a strong baseline.

The results for the XNLI task (depicted in Figure~\ref{fig:XNLI_transfer}) exhibit a different pattern.
For XNLI, the sub-networks found for a language often perform well for other languages, hinting at significant language-neutral components in these sub-networks. At times, these transferred sub-networks even exceed the performance of the source language sub-networks.
For example, the French and English winning tickets are also winning tickets for all other languages we examined.
The Spanish, Russian, and German sub-networks also transfer well to most other languages.
However, the transfer performance of Urdu\footnote{For XNLI, the sparsest Urdu winning ticket was at $s=30\%$. For consistency, we used a 50\% sparse sub-network for Urdu in these experiments, but the 30\% sparse sub-network does not transfer well to other languages either.} and Swahili sub-networks are worse than other languages, possibly because these two languages were under-represented during mBERT's pre-training, and so contributed less to the model's final parameters compared to other languages\footnote{Urdu and Swahili have less than 200k Wikipedia articles while the rest are on the scale of millions of articles:  \href{https://meta.wikimedia.org/wiki/List_of_Wikipedias}{https://meta.wikimedia.org/wiki/List\_of\_Wikipedias}}.
%
We also note that Arabic's winning ticket outperforms mBERT by $\sim$1.5 points for Swahili and Urdu. 
As both Urdu and Swahili have been historically influenced by Arabic~\cite{spear2000early, versteegh2014arabic}, they may benefit from Arabic being a high resource language in the pre-training corpus.

\begin{figure}
\centering
\includegraphics[width=0.9\columnwidth]{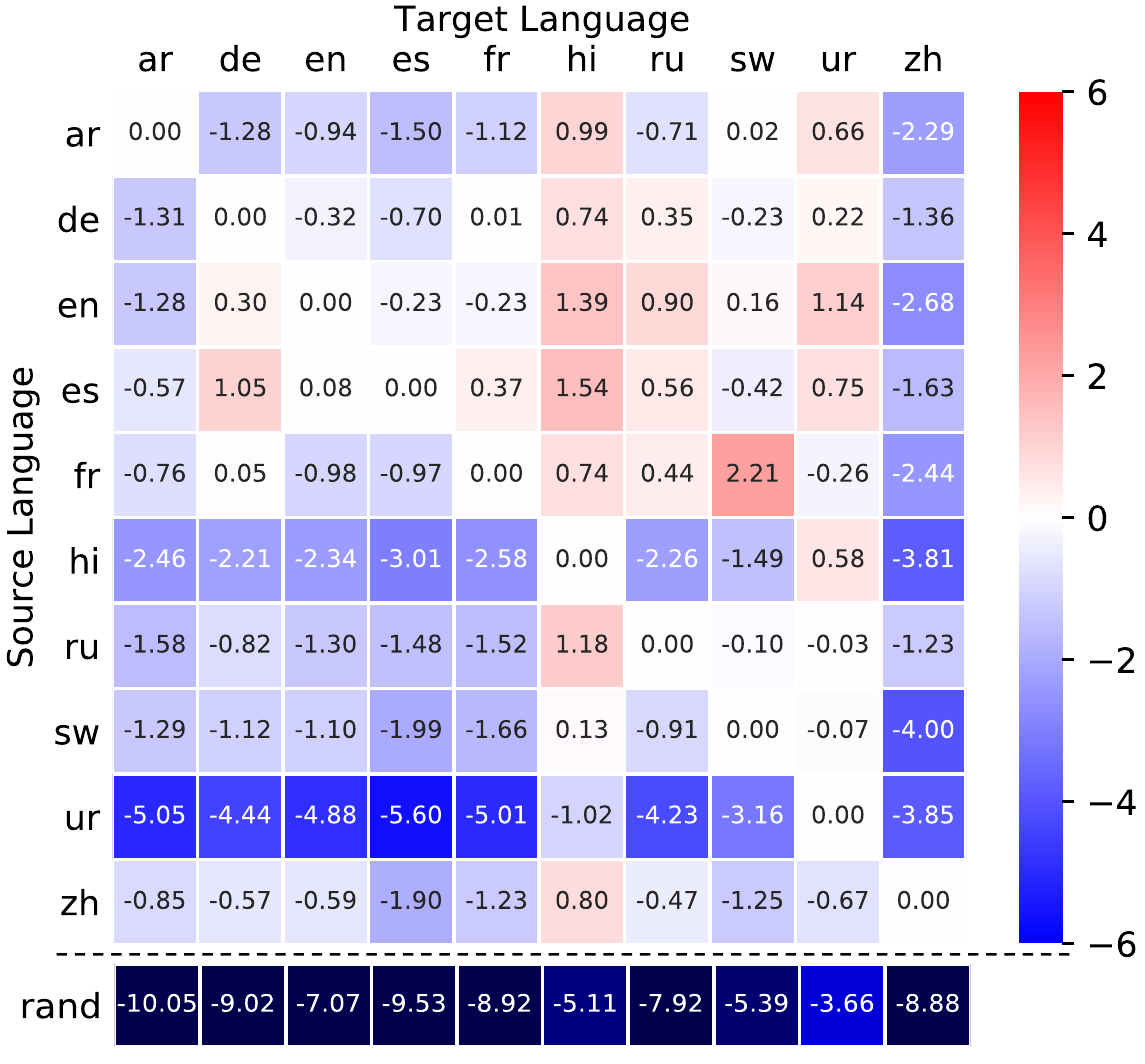}
\caption{The performance difference of transferring
winning tickets across languages (50\% sparsity) for the XNLI task on the mT5 model.}
\vspace{-4mm}
\label{fig:mt5_results}
\end{figure}

Interestingly, the parameter overlaps of all the language pairs do not always predict their cross-lingual transfer performance. 
For MLM, the lower relative degree of sub-network overlap between languages does align with the observed performance drop when transferring winning tickets between languages~(Figure~\ref{fig:MLM_transfer}). Similarly, an increased overlap is observed among XNLI sub-networks, corresponding to improved transfer performance between languages. When looking at the NER sub-networks, however, which have the highest pairwise overlap, we do not observe successful transfer results~(Figure~\ref{fig:NER_transfer}). We conjecture that transfer for NER may require specialized knowledge about the entities likely to be discussed in a particular language (\eg, Chinese Wikipedia articles in the NER dataset may contain more information about Chinese public and historical figures). Consequently, even if a high overlap is observed, the non-overlapping parameters in these sub-networks are crucial for successful task performance\footnote{Further analysis is in Appendix~\ref{sec:cross-task}.}.

To investigate to what extent these observations generalize to other models, we repeat the same experiments for the mT5 model~\citep{xue2021MT5}.\footnote{\href{https://huggingface.co/google/mt5-base}{https://huggingface.co/google/mt5-base}}
This model is a multilingual text-to-text transformer with 12 layers and 580 million parameters and is trained on a multilingual variant of the C4 dataset~(mC4; \citealp{raffel2020C4}) covering 101 languages.
We formulate the XNLI task into a text-to-text format similar to the mT5 paper by generating the label text from the concatenation of the premise and hypothesis.
Figure~\ref{fig:mt5_results} shows the cross-lingual transfer performance for the XNLI task at 50\% sparsity level.\footnote{We note that winning tickets for these languages are not found for this sparsity level, but chose to maintain consistency with the mBERT experiments.}
For most cases, the transfer performance drop is relatively small, and, as with mBERT, similar languages such as English, Spanish, French, and German transfer well to most of the other languages, suggesting that language-neutral parameters are a common phenomenon in different MultiLMs.

\begin{figure}
    \centering
    \centering
    \subfloat[NER]{
        \label{fig:ner_perf_drop}
        {\includegraphics[width=0.85\columnwidth]{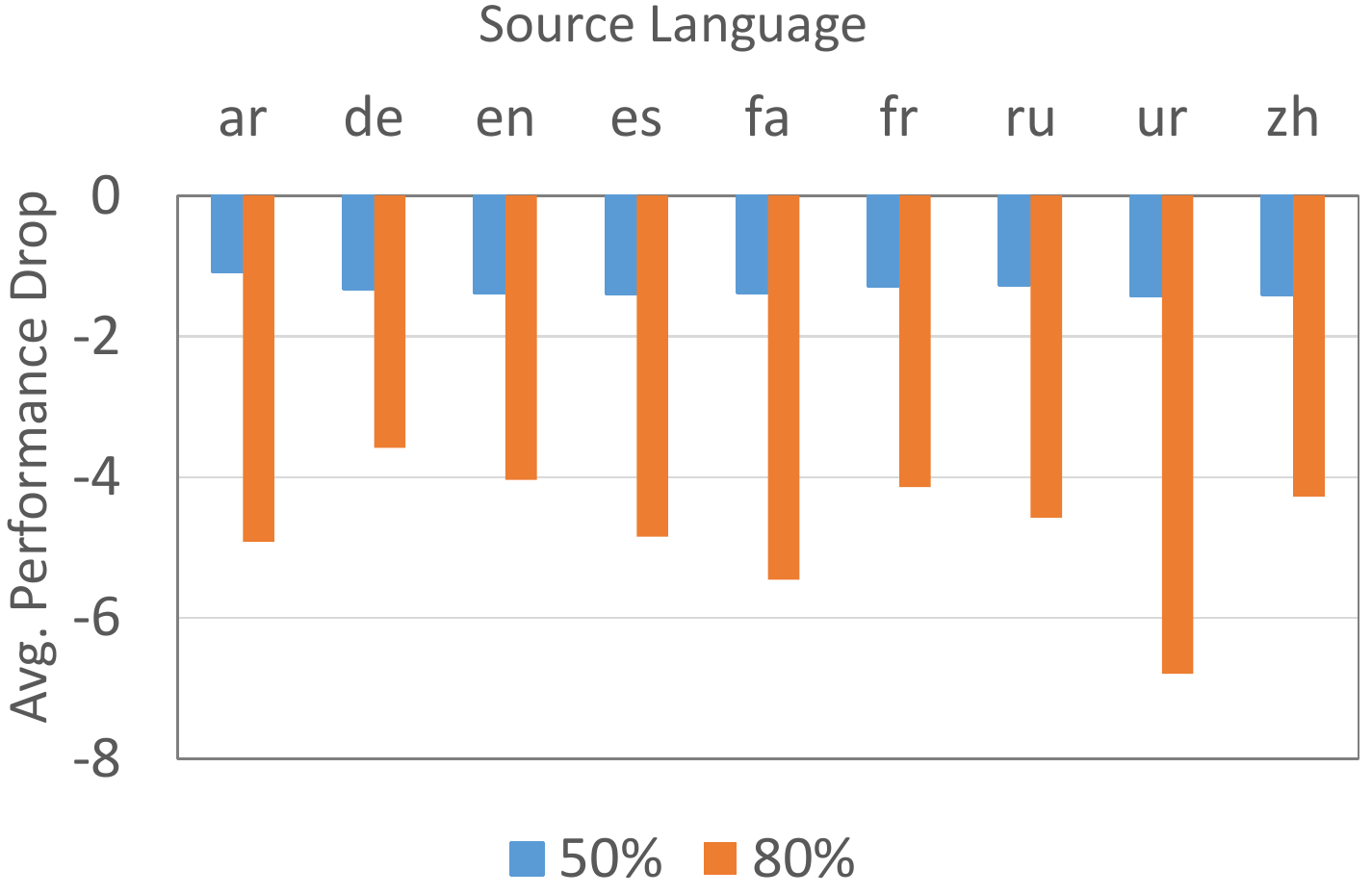}}
    }
    
    \centering
    \subfloat[XNLI]{
        \label{fig:xnli_perf_drop}
        {\includegraphics[width=0.85\columnwidth]{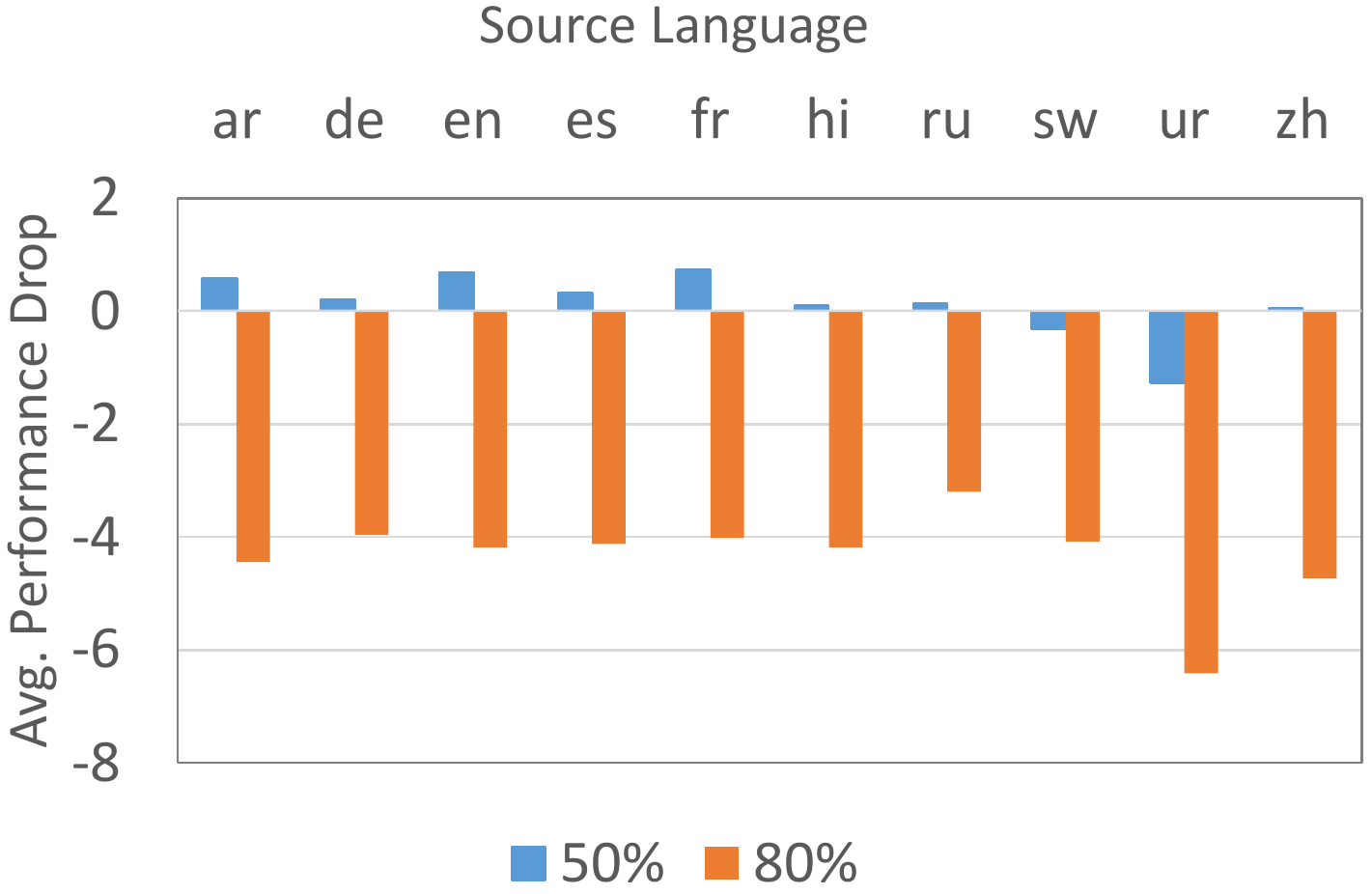}}
    }
    \caption{Average cross-transfer performance drop for sub-networks with sparsity levels 50\% and 80\%. For each source language and task, the relative average is computed across the other languages and the same task.}
    \label{fig:avg_perf_drop}
    \vspace{-4mm}
\end{figure}

\paragraph{Impact of Sub-network Density.} To investigate the effect of sparsity on the retainment of language-neutral components, we compare the cross-lingual transferability of mBERT sub-networks at 50\% and 80\% sparsity levels. 
Figure~\ref{fig:avg_perf_drop} shows the average of relative transfer performance drop\footnote{$\frac{1}{\vert L \vert - 1}\sum_{t \in L \setminus s} \frac{a(s, t) - a(t, t)}{a(t, t)}$ where $s$ and $t$ are source and target languages and $L$ is the set of languages for each task.} 
per language for the NER and XNLI tasks. Each bar represents an average performance drop after retraining (and evaluating) the sub-network for the source language on all the target languages, individually.
As we increase the sparsity level of a sub-network, its cross-lingual transferability degrades considerably (\ie, the \textit{relative} performance drop \textit{increases}), indicating that there were language-neutral parameters at 50\% sparsity level facilitating the cross-lingual transfer that were pruned in the 80\% sparse sub-network.
As we decrease the model's capacity by pruning more parameters, the model relies more on language- and task-specific parameters than those that may facilitate the cross-lingual transfer.

\begin{figure*}
    \centering
    \subfloat{
        {\includegraphics[width=0.75\textwidth]{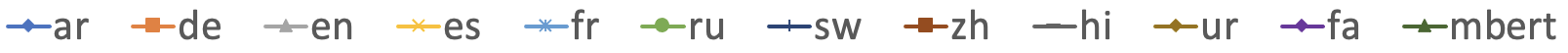}}
    }
    
    \addtocounter{subfigure}{-1}
    
     \centering
    \subfloat[MLM]{
        \label{fig:mlm_translation}
        {\includegraphics[width=0.32\textwidth]{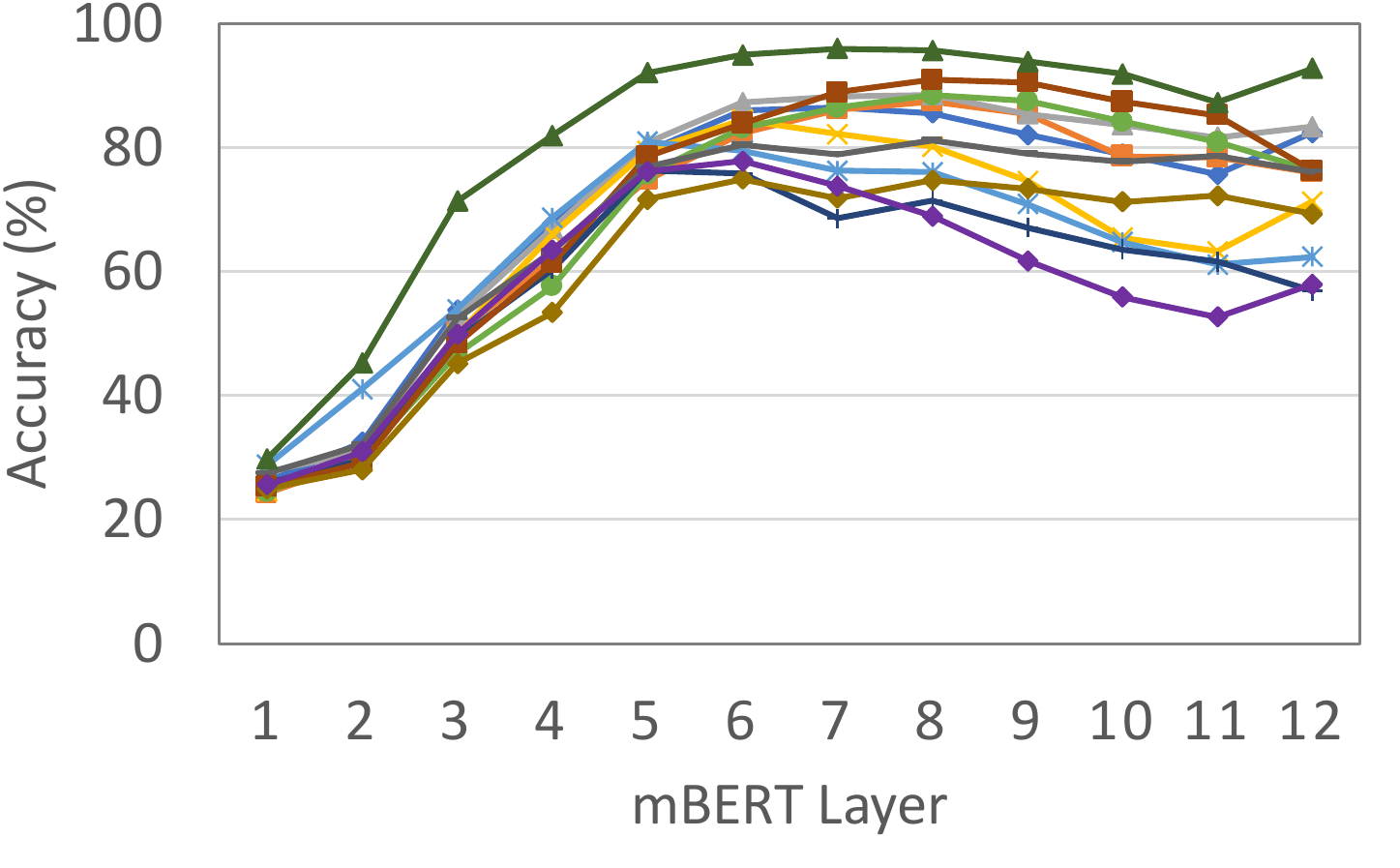}}
    }
    \subfloat[NER]{
        \label{fig:ner_translation}
        {\includegraphics[width=0.32\textwidth]{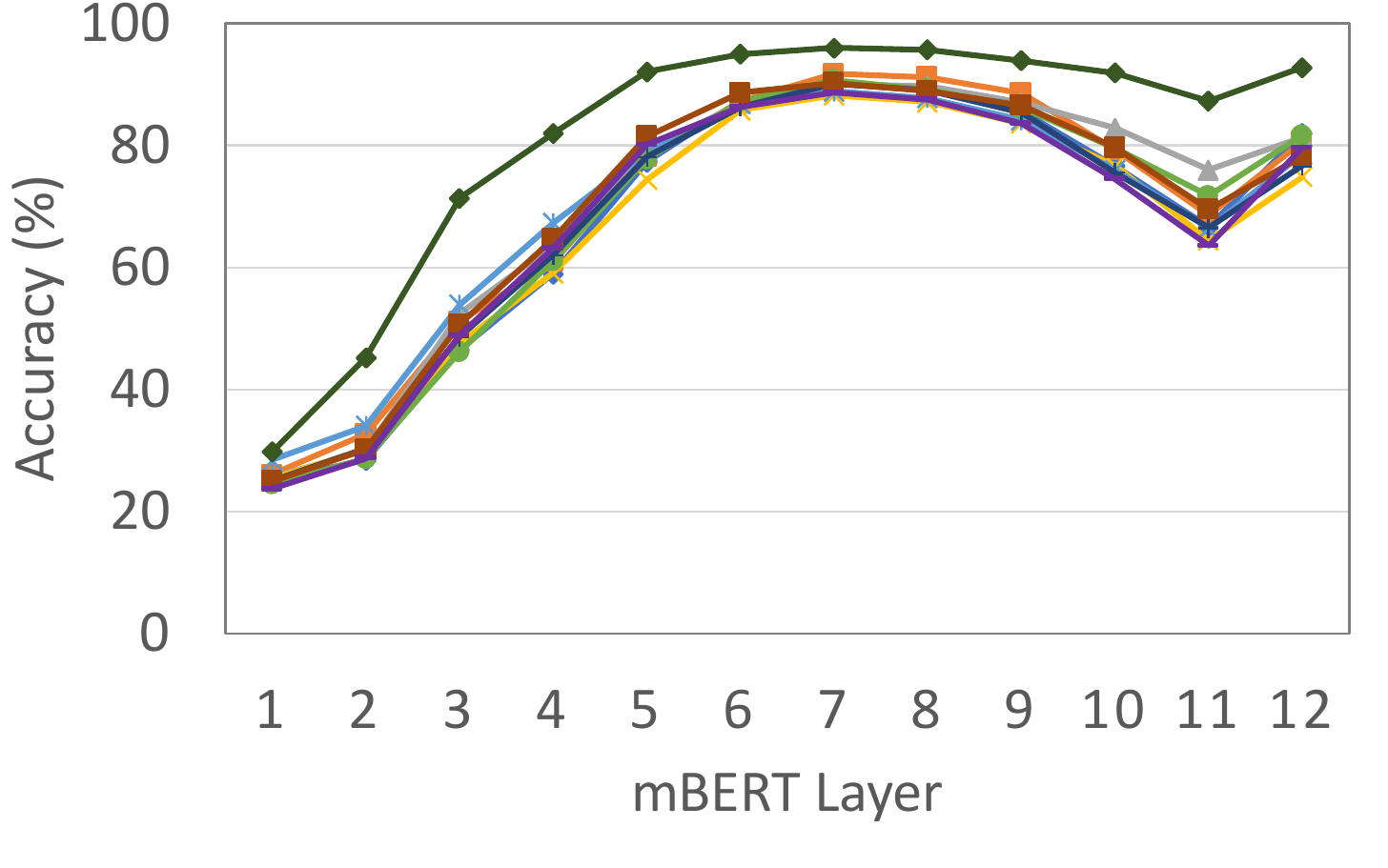}}
    }
    \subfloat[XNLI]{
        \label{fig:xnli_translation}
        {\includegraphics[width=0.32\textwidth]{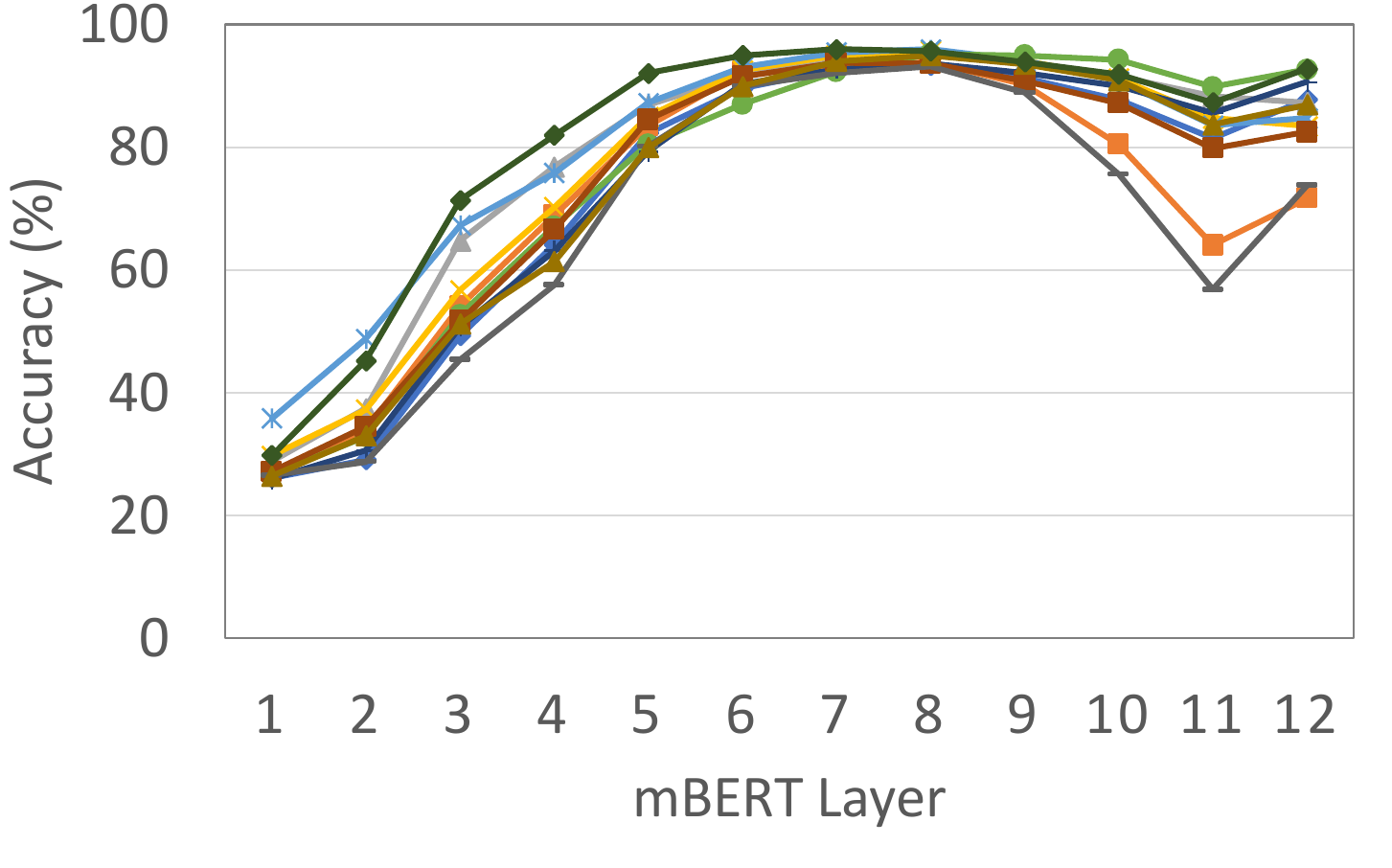}}
    }
        \caption{Parallel sentence retrieval accuracy on English-to-French translation using winning tickets.
        We use the average of the contextual embeddings of each sentence as the retrieval inputs.}
        \label{fig:translation_accuracy}
    \vspace{-4mm}
\end{figure*}

\paragraph{Parallel Sentence Retrieval.}

To further assess the language neutrality of mBERT, we compare the behavior of different sub-networks in a zero-shot setting where they are used as feature extractors. 
We evaluate each sub-network's sentence retrieval accuracy~\cite{pires2019multilingual} on the English-to-French translation test set from the WMT14 dataset~(3003 sentence pairs).\footnote{\href{https://huggingface.co/datasets/wmt14}{https://huggingface.co/datasets/wmt14}}
This task aims to find sentence pairs from two corpora in two different languages, where the two sentences of each pair are corresponding translations of one another. 
In these experiments, we use the parallel retrieval implementation from LASER\footnote{\href{https://github.com/facebookresearch/LASER}{https://github.com/facebookresearch/LASER}} with the margin-based scoring function from~\cite{artext2019margin}.\footnote{The definition of the function is available in Appendix~\ref{sec:scoring_func}.} 
We use the average of the token embeddings for each sentence (encoded using the sub-networks) as the retrieval inputs.




We observe that both the similarity across the sub-networks (Figure~\ref{fig:mlm_sparsity_overlap}) and the top-5 retrieval accuracy (Figure~\ref{fig:translation_accuracy}) increase as we go into the middle layers of mBERT. 
XNLI's winning tickets perform better than the other two tasks, and some of these sub-networks even reach the same zero-shot accuracy as the full mBERT model.
One possible explanation is that both the XNLI task and parallel sentence retrieval require a richer semantic interpretation of the text. Hence, XNLI winning tickets are better sub-networks for parallel sentence retrieval as they must capture semantic knowledge to successfully complete the task. 
In the case of NER, the sub-networks of different languages perform similarly across all layers, which is unsurprising given the high amount of overlap between these sub-networks. In a zero-shot setting with no language-specific tuning for the task, these sub-networks are basically identical.
For MLM, the sub-networks perform similarly up to the middle layers, where they begin to diverge, supporting the hypothesis that higher layers in the MLM sub-networks are dedicated toward representing the task, which is language-specific for MLM. 

\paragraph{Multilingual Sub-networks.}
Observing the benefit of transferring sub-networks from high-resource languages to lower resource ones (\eg, Arabic to Urdu and Swahili for NER), we study the impact of combining datasets from multiple languages to discover multilingual winning tickets.
In this set of experiments, we combined English, Farsi, French, and Chinese datasets together to prune mBERT for the NER task. For parity, we use an identically-sized combined training dataset with 20,000 samples (the same as other experiments on this task), keeping only 25\% of each of each language's dataset.
Table~\ref{table:combined_datasets} shows the transfer results of the resulting sub-network at 50\% sparsity level.
For three out of five languages that we evaluated transfer performance on, the sub-network extracted using the combination of languages outperforms the sub-networks found using each training language separately. The inclusion of multiple languages in the corpus may encourage recovering more language-neutral sub-networks than discovering sub-networks using only a single language.

\begin{table}
\centering
\resizebox{\columnwidth}{!}{
\begin{tabular}{ l c c c c c}
\toprule
\textbf{Model /} & \multicolumn{5}{c}{\textbf{Target Language}} \\
\textbf{Sub-network}  & \textbf{ar} & \textbf{de} & \textbf{es} & \textbf{ru}  & \textbf{ur} \\
\midrule
\textbf{mBERT}  & 88.69 & 89.08 & 91.11 & 89.22 & 95.52\\
\midrule
\textbf{$f^*$ } & 88.5  & 89.01 & 91.08 & 89.03 & 96.44\\
\midrule
\textbf{en} & 87.23 & 87.52	& 90.44	& 88.47	& 93.93\\
\textbf{fa} & 87.56 & 87.57 & 90.18 & 88.34 & 94.10 \\
\textbf{fr} & 87.41	 & \textbf{87.83} & 90.56	 & \textbf{88.50}	 & 94.42\\
\textbf{zh} &  87.01 & 87.67 & 90.11 & 88.21 & 93.65\\
\textbf{en-fa-fr-zh} & \textbf{87.64} & 87.81 & \textbf{90.77} & 88.34 & \textbf{94.59} \\

\midrule
\end{tabular}
}
\caption{Performance (F1-score) of winning tickets at 50\% sparsity level for combining training datasets of en, fa, fr, and zh languages for the NER task.}
\vspace{-4mm}
\label{table:combined_datasets}
\end{table}
\section{Conclusion}
\label{sec:conclusion}


While multilingual pre-trained language models 
have shown impressive performance across languages, the role of language neutrality in achieving such a performance is not well understood.
In this work, we analyze the language-neutrality of multilingual models by investigating the overlap between language-encoding sub-networks of these models.
Using mBERT as a foundation, we expose the extent of its language neutrality by employing the lottery ticket hypothesis and comparing the sub-networks within mBERT obtained for various languages and tasks.
We show that such sub-networks achieve high performance after being transferred across tasks and languages and are similar to one another. However, at higher levels of sparsity, this transferability evaporates. Our results suggest that multilingual language models include two separate language-neutral and language-specific components, with the former playing a more prominent role in cross-lingual transfer performance.
\section*{Limitations}
In this work, we use network parameter overlap to measure similarity between sub-networks discovered for different languages.
However, while cross-lingual transfer performance coarsely tracks with sub-network overlap, these results may not hold at a fine-grained level. While high sub-network overlap is an indicator that a sub-network will effectively transfer to a target language, small relative differences in cross-lingual performance across languages do not correlate strongly with overlap. Consequently, absolute overlap may be limited as an analogue for identifying language-neutral components of multilingual models.
Another limitation is that we only use the lottery ticket hypothesis as a method for discovering language-specific sub-networks, while other pruning and masking methods may provide complementary or competing insights.
Finally, due to computational limitations, we only apply our work to three tasks and eleven languages. While our selection is diverse, new insights may emerge from a larger cross-section.
\section*{Acknowledgements}
The authors thank the anonymous reviewers for their valuable comments and feedback.
We also thank the members of LSIR and NLP labs at EPFL for their feedback and support. 
Antoine Bosselut gratefully acknowledges the support of Innosuisse under PFFS-21-29, the EPFL Science Seed Fund, the EPFL Center for Imaging, Sony Group Corporation, and the Allen Institute for AI.

\bibliography{gen-abbrev,dblp,misc}
\bibliographystyle{acl_natbib}

\appendix


\section{Additional Lottery Ticket Results}
\label{sec:more_result_on_lth}

Tables~\ref{table:mlm_result_all}, \ref{table:xnli_result_all}, and \ref{table:ner_result_all} show the performance of winning tickets across languages for the MLM, XNLI, and NER tasks respectively.
The absolute performance values of all three tasks on mBERT model are depicted in Tables~\ref{table:mlm_result_all}, \ref{table:xnli_result_all}, and \ref{table:ner_result_all}.
Table~\ref{table:mt5_xnli_result} shows the absolute performance for the XNLI task on the mT5 model.

\begin{table*}[t]
\begin{center}

\resizebox{\textwidth}{!}{\begin{tabular}{l|rrrr rrrr rrrr}
\toprule
\multicolumn{1}{c}{} & \multicolumn{4}{c}{\textbf{MLM (Perplexity)}} & \multicolumn{4}{c}{\textbf{NER (F1)}} & \multicolumn{4}{c}{\textbf{XNLI (Accuracy)}}\\
\cmidrule(lr){2-5}                   \cmidrule(lr){6-9}     \cmidrule(lr){10-13}      
 \multicolumn{1}{c}{} & \textbf{mBERT} & $f^*$ & $s$~(\%) &  \multicolumn{1}{r}{\textbf{rand}} & \textbf{mBERT} & $f^*$ & $s$~(\%) & \multicolumn{1}{r}{\textbf{rand}} & \textbf{mBERT} & $f^*$ & $s$~(\%) & \textbf{rand}\\
\toprule

ar & 3.5247 & 3.5546 & 50 & 6.4001  & 88.64 & 88.50 & 50 & 72.43 & 70.26 & 70.29 & 50 & 61.48 \\
de & 3.5143 & 3.5092 & 50 & 9.6489  & 89.11 & 88.81 & 50 & 75.68 & 77.33 & 77.10 & 50 & 65.98 \\
en & 4.6523 & 4.6347 & 50 & 10.9182 & 83.47 & 83.60 & 50 & 68.05 & 82.16 & 82.14 & 50 & 74.49 \\
es & 3.6775 & 3.6712 & 50 & 8.3084  & 91.11 & 91.08 & 50 & 79.44 & 78.80 & 78.93 & 60 & 67.67  \\
fa & 3.8033 & 3.8038 & 50 & 7.3315  & 92.33 & 92.10 & 60 & 75.66 &    -  &   -   & -  &   -   \\
fr & 3.1151 & 3.0936 & 50 & 6.8504  & 90.51 & 90.31 & 50 & 77.61 & 78.00 & 77.61 & 50 & 69.11 \\
hi & 2.8728 & 2.8757 & 50 & 5.1345  &    -  &   -   & -  &    -  & 68.48 & 68.04 & 50 & 58.75 \\
ru & 2.5927 & 2.5907 & 50 & 6.1112  & 89.39 & 89.03 & 50 & 74.91 & 72.98 & 72.73 & 60 & 62.20  \\
sw & 2.5001 & 2.4657 & 50 & 4.292   &    -  &   -   & -  &    -  & 66.28 & 66.14 & 50 & 59.55 \\
ur & 2.8624 & 2.8638 & 50 & 5.0286  & 95.31 & 95.94 & 60 & 84.33 & 63.06 & 63.13 & 30 & 58.75 \\
zh & 3.6096 & 3.5754 & 50 & 8.4811  & 79.53 & 79.33 & 50 & 57.44 & 76.60 & 76.01 & 50 & 68.23 \\

\midrule
\end{tabular}
}
\caption{Performance of winning tickets~($f^*$) on the XNLI, NER, and MLM tasks at the highest sparsity~($s$) for which iterative pruning finds them.
We also report the performance of a random sub-network of mBERT (\textbf{rand}) at the same sparsity level as the winning ticket.}

\label{table:all_in_result}
\vspace{-4mm}
\end{center}
\end{table*}

\begin{table*}[t]
  \centering
\resizebox{\textwidth}{!}{
\begin{tabular}{ l c c c c c c c c c c c}
\toprule
\textbf{Model /} & \multicolumn{11}{c}{\textbf{Target Language}} \\
\textbf{Sub-network}  & \textbf{ar} & \textbf{de} & \textbf{en} & \textbf{es} & \textbf{fa} & \textbf{fr} & \textbf{hi} & \textbf{ru} & \textbf{sw} & \textbf{ur} & \textbf{zh} \\
\midrule
\textbf{mBERT} & 3.5247 & 3.5143 &  4.6523 & 3.6775 & 3.8033 & 3.1151 & 2.8728 & 2.5927 & 2.5001 &  2.8624 & 3.6096 \\
\midrule
\textbf{ar} & 3.5546 & 3.7510 & 4.7967 & 3.7568 & 4.0791 & 3.2360 & 3.1133 & 2.7075 &2.7315 &3.0873 & 3.7821\\
\textbf{de} & 3.8861 & 3.5092 & 4.9626 & 3.9055 & 4.1539 & 3.3429 & 3.1189 & 2.8173 & 2.7535 & 3.0852 & 3.9186 \\
\textbf{en} & 3.8742 & 3.8270 & 4.6347 & 3.8585 & 4.2320 & 3.3177 & 3.1034  & 2.8129 & 2.7436 & 3.0899 & 3.9407 \\
\textbf{es} & 3.8339 & 3.7929  & 4.8842 & 3.6712 & 4.1942 & 3.2895 & 3.0974 & 2.7693 & 2.7569 & 3.1046 & 3.8816\\
\textbf{fa} & 3.7464 & 3.6334 & 4.7350 & 3.6938 & 3.8038 & 3.1910 & 3.0848 & 2.6704 & 2.7339 & 3.0595 & 3.7303 \\
\textbf{fr} & 3.8722 & 3.8528 & 4.9441 & 3.8526 & 4.1329 & 3.0936 & 3.1116 & 2.8150 & 2.7406 &3.1577 &3.9641 \\
\textbf{hi} & 3.8431 & 3.7554 & 4.8606 & 3.8120 & 4.2017 & 3.2725 & 2.8757 & 2.7432 & 2.7612 & 3.1281 & 3.8329\\
\textbf{ru} & 3.8820 & 3.8683 & 5.0007 & 3.9182 & 4.1466 & 3.3656 & 3.0847 & 2.5907 & 2.7566 & 3.1553 & 3.9728 \\
\textbf{sw} & 3.8650 & 3.7613 & 4.8704 & 3.8184 & 4.2288 & 3.2863 & 3.1022 & 2.7505  & 2.4657  & 3.0666 & 3.8442 \\
\textbf{ur} & 3.8644 & 3.7545 & 4.8701 & 3.8153 & 4.1892 & 3.2778 & 3.0937 & 2.7456 & 2.7575 & 2.8638 & 3.8410 \\
\textbf{zh} & 3.8852 & 3.8804 & 5.0004 & 3.9216 & 4.1565 & 3.3698 & 3.1025 & 2.8302 & 2.7675 & 3.1015 & 3.5754 \\


\midrule
\end{tabular}
}
\caption{Performance~(perplexity) of winning tickets at $50\%$ sparsity for the MLM task.}
\label{table:mlm_result_all}
\end{table*}
\begin{table*}[t]
\footnotesize
\centering
\begin{tabular}{ l c c c c c c c c c c }
\toprule
\textbf{Model /} & \multicolumn{10}{c}{\textbf{Target Language}} \\
\textbf{Sub-network}  & \textbf{ar} & \textbf{de} & \textbf{en} & \textbf{es}  & \textbf{fr} & \textbf{hi} & \textbf{ru} & \textbf{sw} & \textbf{ur} & \textbf{zh} \\
\midrule
\textbf{mBERT}  & 70.26 & 77.33 & 82.16 & 78.80 & 78.00 & 68.48 & 72.98 & 66.28 & 63.06 & 76.60\\
\midrule
    \textbf{ar} & 70.29 & 76.72 & 82.00 & 79.07 & 77.06 & 67.58 & 72.96 & 67.48 & 64.68 & 76.88\\
    \textbf{de} & 70.22 & 77.10 & 82.44 & 79.09 & 77.73 & 68.11 & 73.35 & 65.40 & 64.03 & 76.50 \\
    \textbf{en} & 71.17 & 77.12 & 82.14 & 79.73 & 78.23 & 67.90 & 74.38 & 66.37 & 64.03 & 76.66\\
    \textbf{es} & 70.57 & 77.42 & 82.43 & 79.15 & 77.36 & 68.06 & 74.31 & 64.97 & 63.61 & 76.98 \\
    \textbf{fr} & 70.52 & 76.96 & 82.80 & 79.56 & 77.61 & 68.39 & 73.31 & 66.77 & 63.85 & 77.31 \\
    \textbf{hi} & 70.94 & 76.24 & 82.54 & 77.96 & 77.15 & 68.04 & 72.93 & 65.78 & 64.07 & 76.50\\
    \textbf{ru} & 70.33 & 76.05 & 82.69 & 79.49 & 77.54 & 68.55 & 73.31 & 65.62 & 63.37 & 76.52 \\
    \textbf{sw} & 69.76 & 76.42 & 81.50 & 77.45 & 76.75 & 67.84 & 72.36 & 66.14 & 63.39 & 76.18 \\
    \textbf{ur} & 69.11 & 75.17 & 80.81 & 77.65 & 75.71 & 66.98 & 72.20 & 65.33 & 62.14 & 75.82\\
    \textbf{zh} & 69.79 & 76.16 & 81.45 & 78.73 & 77.80 & 67.74 & 72.67 & 65.51 & 63.85 & 76.01\\


\midrule
\end{tabular}
\caption{Performance~(accuracy) of winning tickets at $50\%$ sparsity for the XNLI task.}
\label{table:xnli_result_all}
\end{table*}
\begin{table*}[t]
\footnotesize
\centering
\begin{tabular}{ l c c c c c c c c c }
\toprule
\textbf{Model /} & & \multicolumn{8}{c}{\textbf{Target Language}} \\
  \textbf{Sub-network} & \textbf{ar} & \textbf{de} & \textbf{en} & \textbf{es} & \textbf{fa} & \textbf{fr} & \textbf{ru} & \textbf{ur} & \textbf{zh}\\
\midrule
\textbf{mBERT} & 88.64 & 89.11 & 83.47 & 91.11 & 92.33 & 90.51 & 89.39 & 95.31 & 79.53\\
\midrule
\textbf{ar} & 88.50 & 87.59 & 82.92 & 90.39 & 91.73 & 89.84 & 88.34 & 94.67 & 76.97\\
\textbf{de} & 87.09 & 88.81 & 83.06 & 90.43 & 91.20 & 89.59 & 88.38 & 93.93 & 77.25\\
\textbf{en} & 87.21 & 87.53 & 83.60 & 90.44 & 91.23 & 89.67 & 88.47 & 93.93 & 76.92\\
\textbf{es} & 87.10 & 87.59 & 83.10 & 91.08 & 91.33 & 89.93 & 88.25 & 94.22 & 76.51\\
\textbf{fa} & 87.47 & 87.55 & 82.73 & 90.18 & 92.25 & 89.59 & 88.34 & 94.10 & 76.73\\
\textbf{fr} & 87.33 & 87.81 & 82.85 & 90.56 & 91.40 & 90.31 & 88.50 & 94.42 & 76.72\\
\textbf{ru} & 87.26 & 87.84 & 82.99 & 90.43 & 91.41 & 89.75 & 89.03 & 94.35 & 76.96\\
\textbf{ur} & 86.81 & 87.35 & 82.59 & 90.08 & 91.35 & 89.50 & 87.94 & 96.44 & 76.73\\
\textbf{zh} & 86.94 & 87.48 & 82.42 & 90.11 & 91.17 & 89.47 & 88.21 & 93.65 & 79.33\\


\midrule
\end{tabular}
\caption{Performance~(F1-score) of winning tickets at $50\%$ sparsity for the NER task.}
\label{table:ner_result_all}
\end{table*}
\begin{table*}[!ht]
\footnotesize
\centering
\begin{tabular}{ l c c c c c c c c c c }
\toprule
\textbf{Model /} & \multicolumn{10}{c}{\textbf{Target Language}} \\
\textbf{Sub-network}  & \textbf{ar} & \textbf{de} & \textbf{en} & \textbf{es}  & \textbf{fr} & \textbf{hi} & \textbf{ru} & \textbf{sw} & \textbf{ur} & \textbf{zh} \\
\midrule
\textbf{mT5} & 74.17 & 78.52 & 83.13 & 79.64 & 80.14 & 69.98 & 75.56 & 70.00 & 63.25 & 75.10\\
\midrule
\textbf{ar} & 71.25 & 74.43 & 79.69 & 76.14 & 75.47 & 67.18 & 73.51 & 67.24 & 61.14 & 69.15\\
\textbf{de} & 69.94 & 75.71 & 80.31 & 76.94 & 76.60 & 66.93 & 74.57 & 67.00 & 60.70 & 70.09 \\
\textbf{en} & 69.97 & 76.02 & 80.64 & 77.41 & 76.36 & 67.58 & 75.12 & 67.38 & 61.62 & 68.76\\
\textbf{es} & 70.68 & 76.76 & 80.72 & 77.64 & 76.96 & 67.73 & 74.78 & 66.80 & 61.22 & 69.82 \\
\textbf{fr} & 70.49 & 75.76 & 79.66 & 76.67 & 76.59 & 66.93 & 74.66 & 69.43 & 60.21 & 69.00 \\
\textbf{hi} & 68.79 & 73.50 & 78.29 & 74.63 & 74.01 & 66.19 & 71.96 & 65.74 & 61.06 & 67.64\\
\textbf{ru} & 69.67 & 74.89 & 79.33 & 76.16 & 75.07 & 67.37 & 74.22 & 67.13 & 60.44 & 70.22 \\
\textbf{sw} & 69.96 & 74.60 & 79.53 & 75.65 & 74.93 & 66.32 & 73.31 & 67.22 & 60.41 & 67.45 \\
\textbf{ur} & 66.20 & 71.28 & 75.76 & 72.04 & 71.58 & 65.17 & 69.99 & 64.07 & 60.48 & 67.60\\
\textbf{zh} & 70.40 & 75.15 & 80.05 & 75.74 & 75.36 & 66.99 & 73.75 & 65.98 & 59.80 & 71.44\\
\textbf{rand} & 61.20 & 66.70 & 73.57 & 68.11 & 67.67 & 61.08 & 66.30 & 61.84 & 56.82 & 62.57\\

\midrule
\end{tabular}
\caption{Performance~(accuracy) of at $50\%$ sparsity for the XNLI task on the mT5 model.}
\label{table:mt5_xnli_result}
\end{table*}

\begin{figure*}
    \centering
    \subfloat[Encoder]{
        \label{fig:mt5_encoder_network_overlap}
        {\includegraphics[width=0.50\textwidth]{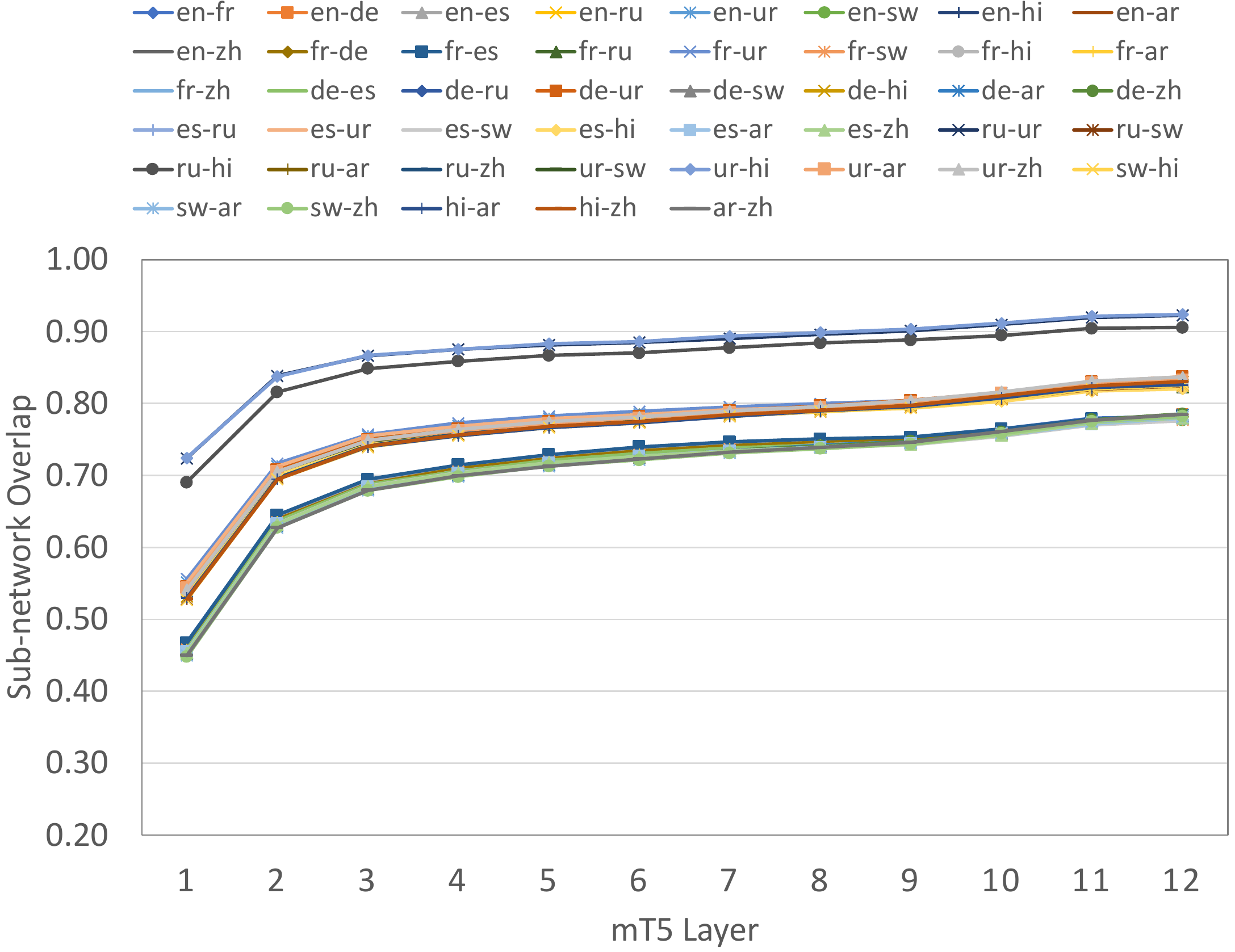}}
    }
    \centering
    \subfloat[Decoder]{
        \label{fig:mt5_decoder_network_overlap}
        {\includegraphics[width=0.50\textwidth]{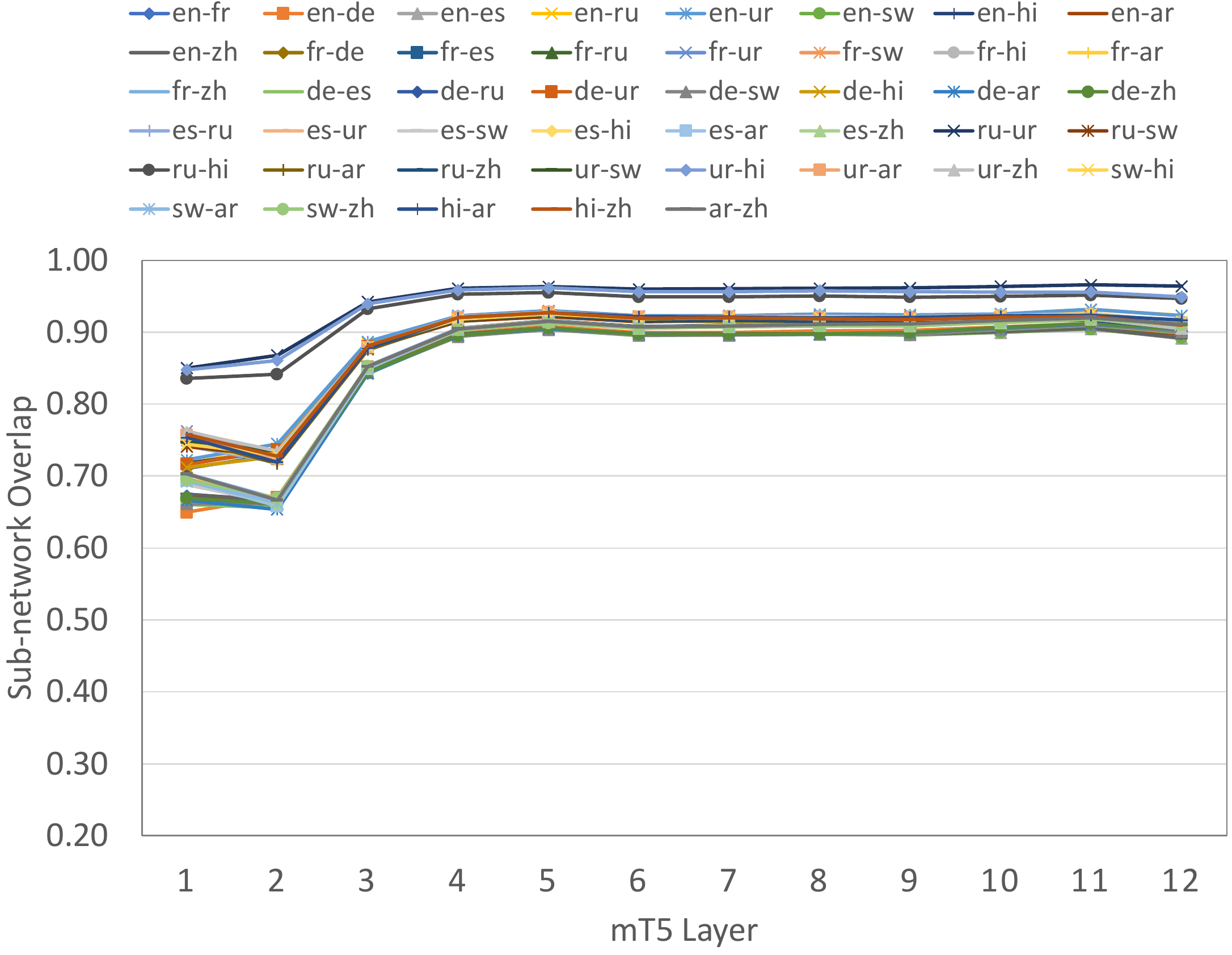}}
    }
    \caption{Sparsity pattern overlap between pruned sub-networks across different layers at 50\% sparsity level for the XNLI task on the mT5 model.}
    \label{fig:mt5_network_overlap}
    \vspace{-4mm}
\end{figure*}

\section{Sub-network Overlap}
\label{sec:subnet_overlap}

In Figure~\ref{fig:mlm_sparsity_overlap}, the parameter overlap plot for the MLM task, we note three different clusters. We observe that these clusters correspond to language pairs between two different groups of languages.
The first set includes English, French, German, Russian, and Chinese and the second set includes the rest of the languages.
The first cluster~(at the bottom) shows the overlaps between the languages of the first set, the third cluster~(on top) includes the overlaps among languages of the second set. The second cluster~(in the middle) covers the overlaps among languages from the first and the second set.

\section{Disentangling Task Neutrality}
\label{sec:cross-task}

Given the significant role of the language-neutral components of the winning tickets, we want to investigate to what extent these components are task-specific.
We evaluate the similarity of winning tickets across tasks for a given language to identify whether the winning ticket for a given language and task can be transferred successfully to other tasks for the same language.
A successful transfer shows that the language-neutral component is not only task-specific but rather a common space across tasks.
Similar to the previous section, we first identify winning tickets for each task-language pair and then train each sub-network on other tasks using the data from the same language.
We again use sub-networks with 50\% sparsity.
We limit the tested languages to those whose sub-networks are present for all three tasks: English~(en), French~(fr), and Chinese~(zh).

\begin{figure}
     \centering
    \subfloat[Target task: MLM ]{
        \label{fig:mlm_cross_task}
        {\includegraphics[width=0.9\columnwidth]{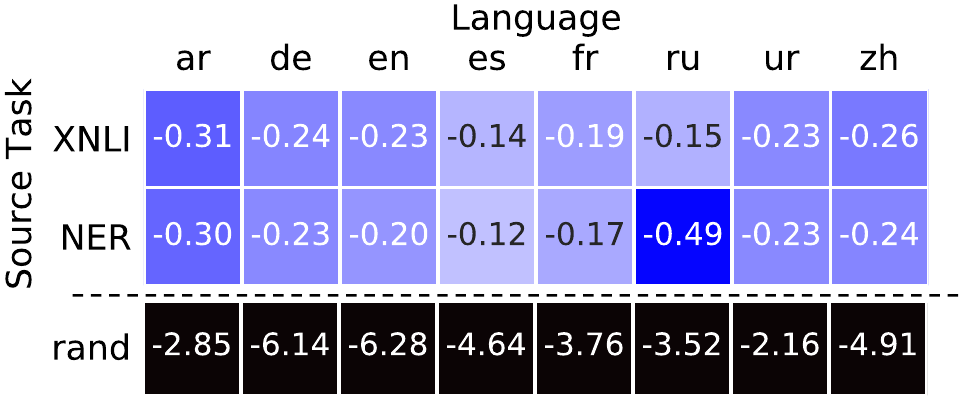}}
    }
    
    \centering
    \subfloat[Target task: NER ]{
        \label{fig:ner_cross_task}
        {\includegraphics[width=0.9\columnwidth]{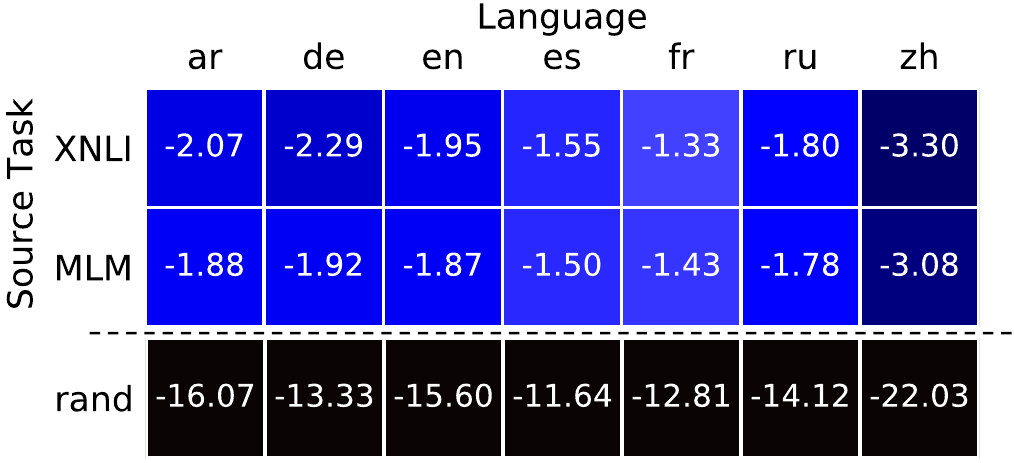}}
    }
    
    \centering
    \subfloat[Target task: XNLI]{
        \label{fig:xnli_cross_task}
        {\includegraphics[width=0.9\columnwidth]{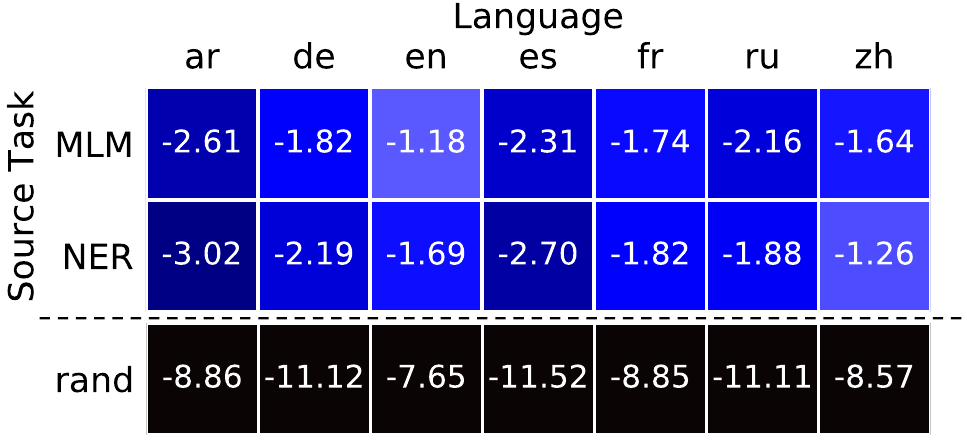}}
    }
    
    \caption{The performance of transferring winning tickets between tasks~(50\% sparsity).
        Each cell shows the difference in performance between the transferred sub-network and the performance of the sub-network discovered for the target task. 
        }
        \label{fig:cross_task_transfer}
    \vspace{-4mm}
\end{figure}

We report cross-task performance in Figure~\ref{fig:cross_task_transfer} for the MLM~(\ref{fig:mlm_cross_task}), NER~(\ref{fig:ner_cross_task}), and XNLI~(\ref{fig:xnli_cross_task}) tasks.
Although in none of the cases the transferred sub-network's performance matches the original sub-network's performance, the average performance drop is less than 2 points for each task.
This observation suggests that a considerable fraction of parameters are shared across tasks even if a certain number of them remain task-specific.
Contrary to previous observations for the BERT model~\cite{chen2020lottery}, we find that MLM winning tickets do not robustly transfer for training other tasks.
\section{Alternative Masking/Pruning Strategies}

In our experiments, we transfer obtained sparse sub-networks by re-training their original weights on a given task and various languages.
Consequently, it is important that these sub-networks match the test accuracy of the full model when trained in isolation~(\ie, winning tickets in the LTH), to be a good sub-network for the task at hand.
By using such a sub-network, we want to keep task-specific parameters as much as possible.
Any pruning/masking methodology that gives us sub-networks performing at the same level as the full model and also uses enough data to detect the language- and task-specific parameters could be used in our analysis.

To broaden our analysis, we ran additional experiments using two alternative masking/pruning strategies. In the first set of experiments, we followed the pruning approach introduced by~\cite{ansell2022composable}.
This method proposes a sparse fine-tuning method to discover task and language sub-networks that can be composed for cross-lingual transfer.
However, unlike the LTH, which prunes parameters with the lowest magnitudes after fine-tuning, this method prunes the parameters that have the smallest absolute difference from the initial parameters.
Following this method, we prune the base model to obtain 50\% sparse sub-networks~(similar to our previous method) for a selection of languages~(\ie, en, fr, de, and ar).
For the NER and XNLI tasks, we observed drops of $\sim$6\% and $\sim$5\%, respectively compared to the full model~(averaged across three random seeds).
As our goal was to find sub-networks that perform as well as the full model for our analysis, these sub-networks would not be suitable candidates.

We also test the method introduced by~\cite{yilin2021fisher}.
This paper proposes a method to approximate the Fisher information matrix as a measure of the importance of each parameter when constructing sparse masks for a given task.
Using a sufficiently large sample size~(1024 examples), they compute this matrix and prune the parameters that contain the least information about the task.
When we use this method~(again pruning to 50\% sparsity), we find that end task performance drops by $\sim$8\% compared to the full model~(on XNLI and NER).
As a result, this approach also does not find high-quality sub-networks that can be tested for cross-lingual transfer.

\section{Margin-based Scoring}
\label{sec:scoring_func}
For the parallel sentence retrieval task, we use the scoring function from~\cite{artext2019margin}:
\begin{multline*}
s(x, y) = \frac{cos(x, y)}{\sum\limits_{z \in \mathcal{N}_{k}(x)} \frac{cos(x, z)}{2k} + \sum\limits_{z \in \mathcal{N}_{k}(y)} \frac{cos(y, z)}{2k}}
\end{multline*}

\noindent where $x$ and $y$ are two sentence representations and $\mathcal{N}_k(x)$ denotes the $k$ nearest neighbors of $x$ in the other language. 
We use $\mbox{margin}(a, b) = a/b$ as our margin function.


\section{Language Representation Similarity}
\label{sec:cca_representation}

\begin{figure}
    \centering
    \subfloat[PWCCA]{
         \label{fig:pwcca_xnli}
        {\includegraphics[width=0.8\columnwidth]{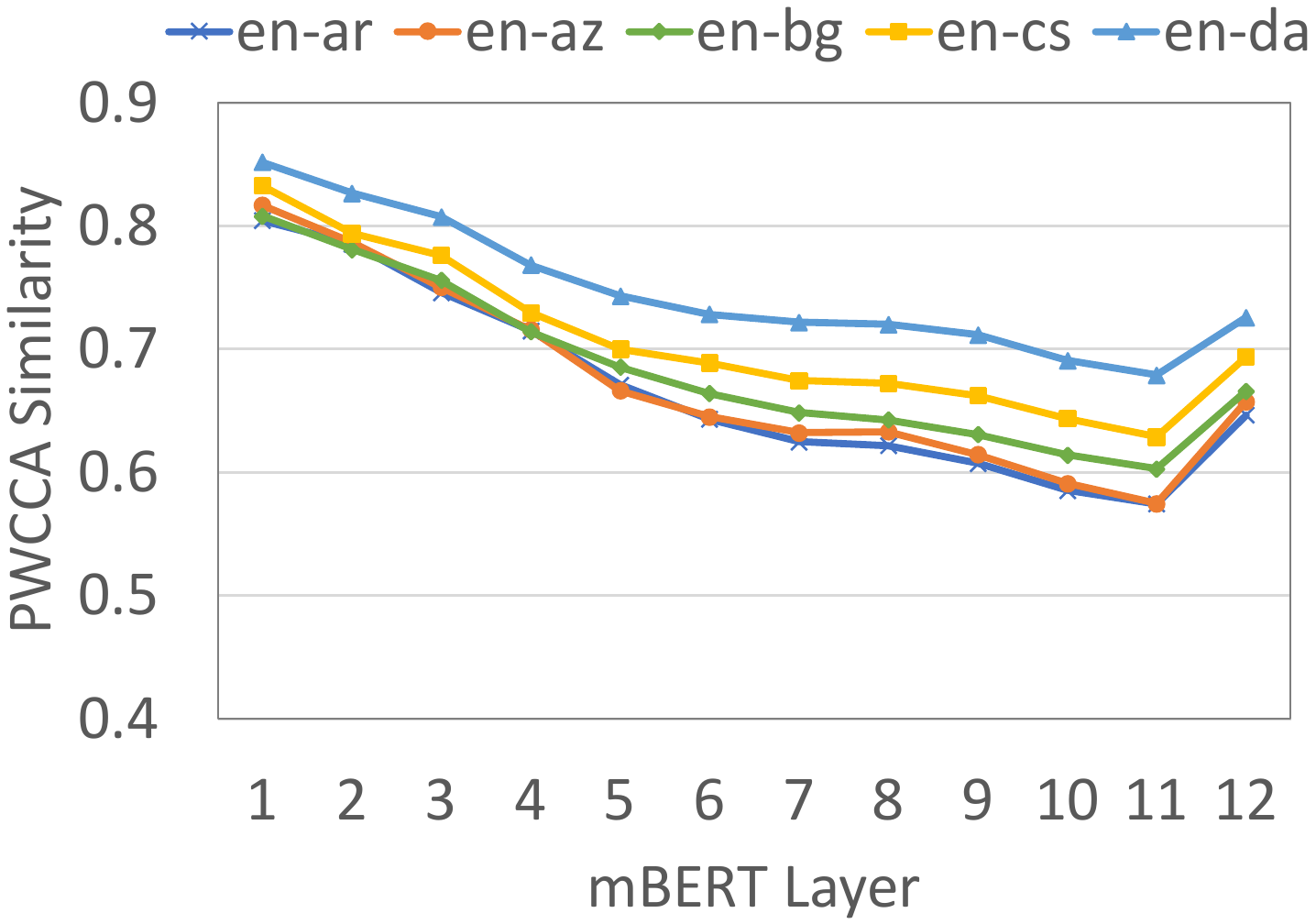}}
    }
    
    \subfloat[SVCCA]{
         \label{fig:svcca_xnli}
        {\includegraphics[width=0.8\columnwidth]{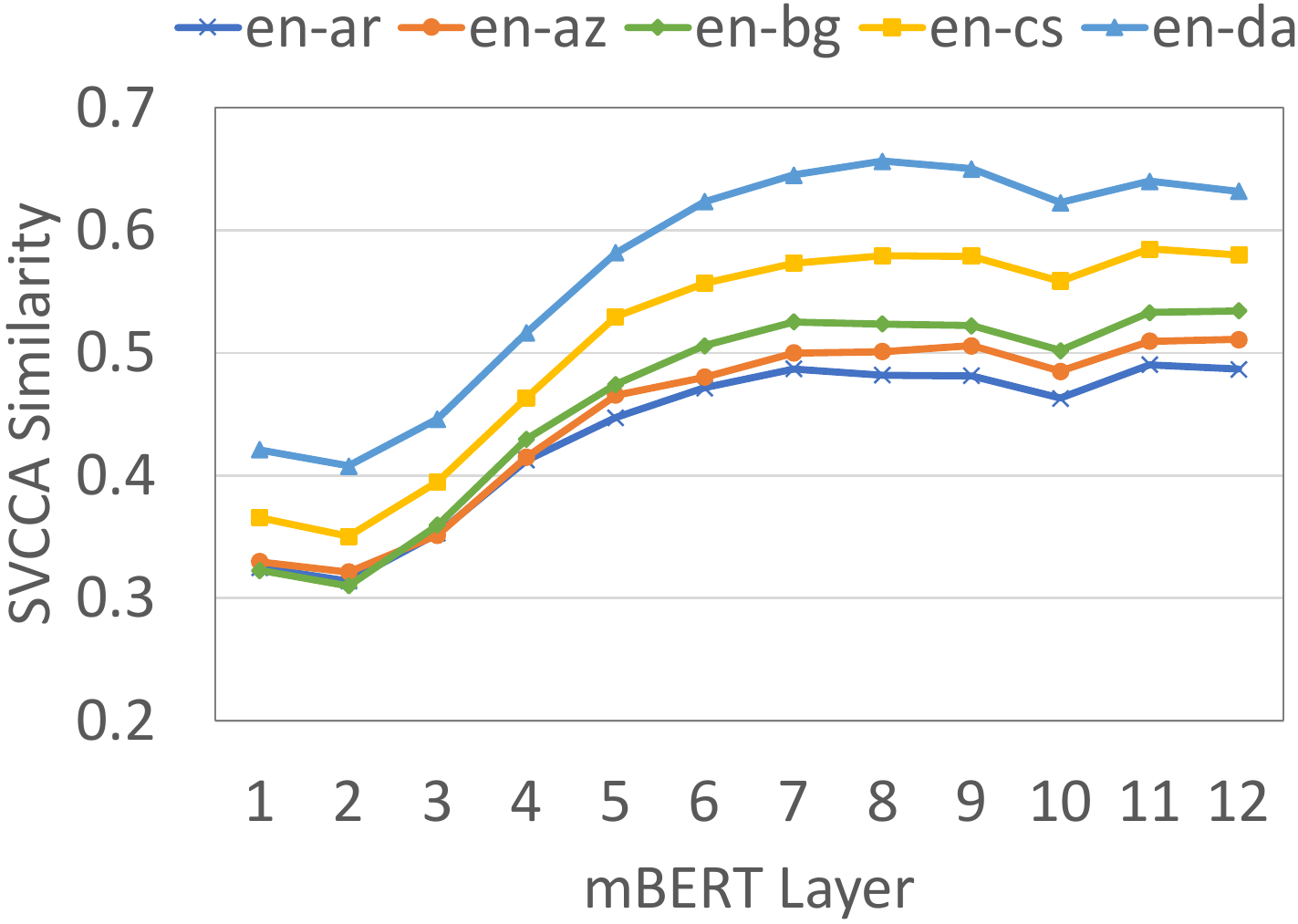}}
    }
    \caption{Similarity scores between CLS representations of English and five other languages using PWCCA and SVCCA.}
    \vspace{-4mm}
\end{figure}

In this section, we discuss a prior work that demonstrated the lack of language neutrality in mBERT by comparing the mBERT's representations for semantically-similar sentences across multiple languages~\cite{singh2019bert}.
They conclude that mBERT partitions the representation space among languages rather than using a shared, interlingual space.
They used projection weighted canonical correlation analysis~(PWCCA; \citealp{morcos2018insights}) to compute the similarity between representations from parallel sentences~(XNLI dataset) for five language pairs.
Figure~\ref{fig:pwcca_xnli} depicts our reproduced version of their results.

As shown in the figure, the CLS representations of various languages are most similar at the shallower layers of mBERT, and their differences grow in deeper layers until the final layer.
We argue that these results contradict mBERT's cross-lingual transfer performance since for almost all of the downstream tasks, the best zero-shot performance is obtained using the representations from middle to deeper layers of mBERT~\cite{conneau2020emerging, chi2020universal, vries2020layers}.
Hence, we expect the representations of deeper layers to be more similar than in shallower layers.

PWCCA is not a reliable tool in this scenario as it is not invariant to orthogonal transformations~\cite{kornblith2019similarity}.
Invariance to orthogonal transformation implies invariance to~(neuron) permutation, which is necessary to accommodate symmetries of neural networks~\cite{kornblith2019similarity}.
Hence, we use singular vector canonical correlation analysis~(SVCCA; \citealp{raghu2017svcca}), which is invariant to orthogonal transformations, to measure the similarities between mBERT's representations in different languages.
SVCCA performs canonical correlation analysis~(CCA) on truncated singular value decomposition of input matrices.

Figure~\ref{fig:svcca_xnli} presents our obtained results for the same five language pairs using SVCCA.
We observe that the similarity of representations increases in deeper layers.
Moreover, the middle and upper layers' representations are more similar across languages than the representations from shallower layers, which is in line with mBERT's cross-lingual transfer performance.
Hence, contrary to prior work~\cite{singh2019bert}, we argue that mBERT maps semantically-similar data points close to each other by learning a common, interlingual space.


\section{Canonical Correlation Analysis}
\label{sec:cca}
Canonical correlation analysis~(CCA) is a statistical tool to identify and measure the associations between two sets of random variables $X$ and $Y$~\cite{hardoon2004cca}.
CCA finds bases for the two input matrices, with the maximum correlation between $X$ and $Y$ projected onto these bases.
For $X \in \mathbb{R}^{d_1 \times n}$, $Y \in \mathbb{R}^{d_2 \times n}$, and $1 \le i \le \mbox{min}(d_1, d_2)$, the $i^{th}$ canonical correlation coefficient $\rho_i$ is given by:
\begin{align*}
\begin{split}
\rho_i = \max_{w^i_X, w^i_Y} \mbox{ corr}(X w^i_X, Y w^i_Y)\\ 
\mbox{subject to } \forall_{j < i} \mbox{  }\: Xw^i_X \: \bot \: Xw^j_X \\
                    \forall_{j < i} \mbox{  }\: Yw^i_Y \: \bot \: Yw^j_Y
\end{split}
\end{align*}
Where $w^i_X \in \mathbb{R}^{d_1}$, $w^i_Y \in \mathbb{R}^{d_2}$ and the constraints enforce orthogonality of the canonical variables.
The average of $\{\rho_1, ..., \rho_m\}$ where $m=\mbox{min}(d_1, d_2)$ is often used as an overall similarity measure:
$$\rho_{CCA} = \frac{\sum_{i=1}^{m} \rho_i}{m}$$

Two of CCA variants, projection weighted canonical correlation analysis~(PWCCA) and singular vector canonical correlation analysis~(SVCCA), are used for analyzing neural network representations because they are invariant to
linear transforms.

To improve the robustness of CCA, SVCCA performs CCA on top of truncated singular value decomposition of $X$ and $Y$.
PWCCA increases the robustness of CCA by using the weighted average of canonical
correlation coefficients:
\[\rho_{PW} = \frac{\sum_{i=1}^{c} \alpha_i\rho_i}{\sum_{i=1} \alpha_i}, \: \: \: \:  \alpha_i = \sum_j \lvert \langle\ h_i, x_j \rangle \rvert\]
where $x_j$ is the $j^{th}$ column of $X$, and $h_i = Xw^i_X$ is the projection of $X$ to the $i^{th}$ canonical coordinate frame.
\section{Computation details}
\label{sec:computation}

Our experiments are executed on a server with 500GB of RAM, $2\times$~16-core Intel Xeon Silver 4216~(2.10GHz) CPUs, and an NVIDIA Quadro RTX 8000 GPU.
For each language, the pruning step for MLM, NER, and XNLI takes around 15, 4.5, and 45 GPU hours, respectively, and fine-tuning takes around 6, 0.5, and 15 GPU hours for MLM, NER, and XNLI, respectively.

\begin{table}[!ht]
\begin{center}
\def\arraystretch{1.2}


\resizebox{\columnwidth}{!}{
\begin{tabular}{ l | c c c}
\hline
 Dataset & MLM & NER & XNLI \\
 \hline
 \# Train Ex. & 1,600,000 & 20,000 & 392,702 \\ 
 \# Valid. Ex. & 30,000 & 10,000 & 2,490\\
 \# Epochs & 1 & 3  & 3\\
 \# Iters/Epoch & 100,000 & 625 & 12,270\\
  Batch Size & 16 & 32 & 32 \\
    Learning Rate & $5 \times 10^{-5}$ & $2 \times 10^{-5}$ & \begin{tabular}{@{}l@{}}
                   mbert: $5 \times 10^{-5}$\\
                   mt5:   $2 \times 10^{-4}$\\
                 \end{tabular} \\
  Eval. Metric & Perplexity & F1 & Accuracy \\
  Optimizer & \multicolumn{3}{ c }{Adam with $\epsilon = 1 \times 10^{-8}$}\\
  \hline
\end{tabular}
}

\caption{Details of pre-training and fine-tuning. Learning rate decays linearly from initial value to zero.}
\vspace{-5mm}
\label{table:train_details}
\end{center}
\end{table}

\end{document}